\documentclass[10pt, a4paper, twocolumn, showabstract]{naverlabseurope}

\usepackage{times}
\usepackage{latexsym}
\usepackage[T1]{fontenc}
\usepackage[utf8]{inputenc}
\usepackage{microtype}
\usepackage{inconsolata}
\usepackage[font=small]{caption}
\usepackage{graphicx}
\usepackage[dvipsnames,table]{xcolor}
\usepackage{tabularx}
\usepackage{booktabs}
\usepackage{multirow}
\usepackage[normalem]{ulem}
\useunder{\uline}{\ul}{}
\usepackage{subcaption}
\usepackage{colortbl}
\usepackage{amsmath}
\usepackage{amssymb}
\usepackage{hyperref}
\usepackage{adjustbox}
\usepackage{pifont}
\usepackage{pgf}

\title{FaST: Feature-aware Sampling and Tuning for\\ Personalized Preference Alignment with Limited Data}
\titlerunning{FaST: Feature-aware Sampling and Tuning for Personalized Preference Alignment with Limited Data}

\authors{Thibaut Thonet$^{1\dagger}$ \authsep Germán Kruszewski$^{1,2}$ \authsep Jos Rozen$^{1}$ \authsep Pierre Erbacher$^{1}$ \authsep Marc Dymetman$^{3}$}
\affiliations{$^{1}$NAVER Labs Europe \quad $^{2}$Universitat Pompeu Fabra \quad $^{3}$Independent Researcher}
\contributions{}
\website{\texttt{\{thibaut.thonet,german.kruszewski,jos.rozen,pierre.erbacher\}@naverlabs.com} \quad \texttt{marc.dymetman@gmail.com}}
\websiteref{}
\definecolor{lightgray}{gray}{0.7}
\newcommand{\ppalli}{PPALLI}

\newcommand{\cmark}{\ding{51}}%
\newcommand{\xmark}{\ding{55}}

\newcommand{\colorvaluegradient}[1]{%
  \pgfmathsetmacro{\val}{int(min(abs(#1)*110,100))}%
  \pgfmathparse{#1 > 0 ? "red!\val!white" : (#1 < 0 ? "blue!\val!white" : "white")}%
  \expandafter\cellcolor\expandafter{\pgfmathresult}{#1}%
}

\begin{abstract}
LLM-powered conversational assistants are often deployed in a one-size-fits-all manner, which fails to accommodate individual user preferences. Recently, LLM personalization~-- tailoring models to align with specific user preferences~-- has gained increasing attention as a way to bridge this gap.
In this work, we specifically focus on a practical yet challenging setting where only a small set of preference annotations can be collected per user~-- a problem we define as Personalized Preference Alignment with Limited Data (\ppalli). To support research in this area, we introduce two datasets~-- DnD and ELIP~-- and benchmark a variety of alignment techniques on them. We further propose FaST, a highly parameter-efficient approach that leverages high-level features automatically discovered from the data, achieving the best overall performance.
\end{abstract}

\begin{document}
\maketitle

\renewcommand{\thefootnote}{\fnsymbol{footnote}}
\footnotetext{$^{\dagger}$Corresponding author:  \texttt{\href{mailto:thibaut.thonet@naverlabs.com}{thibaut.thonet@naverlabs.com}}}
\renewcommand*{\thefootnote}{\arabic{footnote}}
\setcounter{footnote}{0}

\section{Introduction}

Conversational assistants have undergone significant democratization in recent years, resulting in a diverse user base, from the layperson looking for information or entertainment to the IT expert seeking assistance for coding.
Most conversational assistants are provided as a ``one-size-fits-all'' service, and are usually not tailored specifically to individual users and their needs. Because of this, the way the assistant expresses itself may not be entirely suitable to all users. For example, a younger user may prefer simplified answers while a more experienced user is interested in deeper and comprehensive responses. %
We wish to customize the output of Large Language Models (LLMs) and conversational assistants to meet specific criteria, e.g., a user's preferences in a personalized assistant scenario~\citep{jang2023personalized}, a persona's characteristics for role-play and character simulation~\citep{shao_character-llm_2023}, or context-specific principles for Constitutional AI~\citep{Zhan2025}.
Although one could customize the LLMs through prompting by articulating these criteria in words~\citep{zhao2025do}, doing so is often non-trivial. Expressing the criteria accurately can be challenging, leading to potential misunderstandings or imprecise interpretations of the desired outcomes.

An alternative approach, which is the one adopted in this paper, is to construct a preference dataset in which chosen responses reflect the desired criteria.
In particular, we study a preference alignment problem where a \emph{fixed} questionnaire~-- consisting of a set of contexts or questions each paired with several alternative responses~-- is presented to the user targeted for personalization\footnote{In the remainder of the paper, we will refer to the entity targeted for personalization as the ``user'' for the sake of simplicity. This does not imply any loss of generality on the personalization use-case considered.} %
who then picks a single preferred response for each context. In this sense, the questionnaire is considered \emph{user-agnostic}, unlike alternative scenarios where it must be tailored or modified for each specific user. We consider this setting to be a reasonably general and realistic scenario applicable to many practical cases.
We note, however, that it is not practical to require each single user to provide annotations for a large set of contexts, which is what most works in preference alignment rely on~\citep[e.g.,][]{askell2021general, jang2023personalized, zollo2024personalllm}. For this reason, we introduce the problem of \underline{P}ersonalized \underline{P}reference \underline{Al}ignment with \underline{Li}mited Data or \ppalli, which explores how to enable personalization using a small-size questionnaire (i.e., fewer than 100 questions, yielding an equal number of preference tuples). In particular, we focus on training one model per user which offers a clear data confidentiality advantage: users' personal preference data can be kept on their own device where the LLM is directly fine-tuned.%

To facilitate the study of this problem, we introduce and release two new high-quality datasets. The first, DnD, focuses on alignment for the preferences of different characters in a fantasy role-playing setting. The second, ELIP, deals with conversational assistant personalization to respond to open-ended questions in the style preferred by the user.
Moreover, we address the low-data regime problem by introducing FaST (\underline{F}eature-\underline{a}ware \underline{S}ampling and \underline{T}uning), a personalized alignment approach which relies on a highly parameter-efficient reward model. This latter builds on the Compositional Preference Model introduced by~\citet{go2024compositional} and uses high-level interpretable features, automatically discovered from the preference questionnaire presented to users. We evaluate the effectiveness of FaST on the DnD and ELIP datasets\footnote{Our code and the datasets introduced in this work are publicly accessible at \href{https://github.com/naver/fast}{https://github.com/naver/fast}.
} through experiments on both preferred response prediction and personalized generation~-- comparing it with a broad range of preference alignment techniques~-- thereby demonstrating its overall superior performance.

\section{Problem Definition}
\label{sec:problem}

The \ppalli{} problem starts from a small-size questionnaire consisting of various contexts (e.g., instructions to fulfill, questions to answer, or situations to act upon) each paired with several possible, diverse responses to choose from. This set is fixed and presented to all users. Each user selects their preferred response per context, labeled as \textit{chosen}, while others are \textit{rejected}, forming a user-specific preference dataset. This data can then be used to fine-tune an LLM for personalizing its responses. %

This formulation of the LLM personalization task offers the advantage of eliminating the need to generate new responses for each individual user by inferring user preferences from a static questionnaire. In contrast, existing personalized preference datasets commonly assume that either the responses, the contexts, or both are customized for each user \citep{jang2023personalized,kirk2024prism,castricato2024persona}. Such approaches are impractical in realistic scenarios where the system does not have prior knowledge of the user. Alternatively, they would require users to play a more active role in the data collection process~-- for example, by writing and submitting their preferred responses. However, this level of user involvement is typically infeasible, as it is both time-consuming and cognitively demanding for users. %

\section{Feature-aware Sampling and Tuning}

\begin{figure*}[t]
\centering
\includegraphics[width = .995\textwidth]{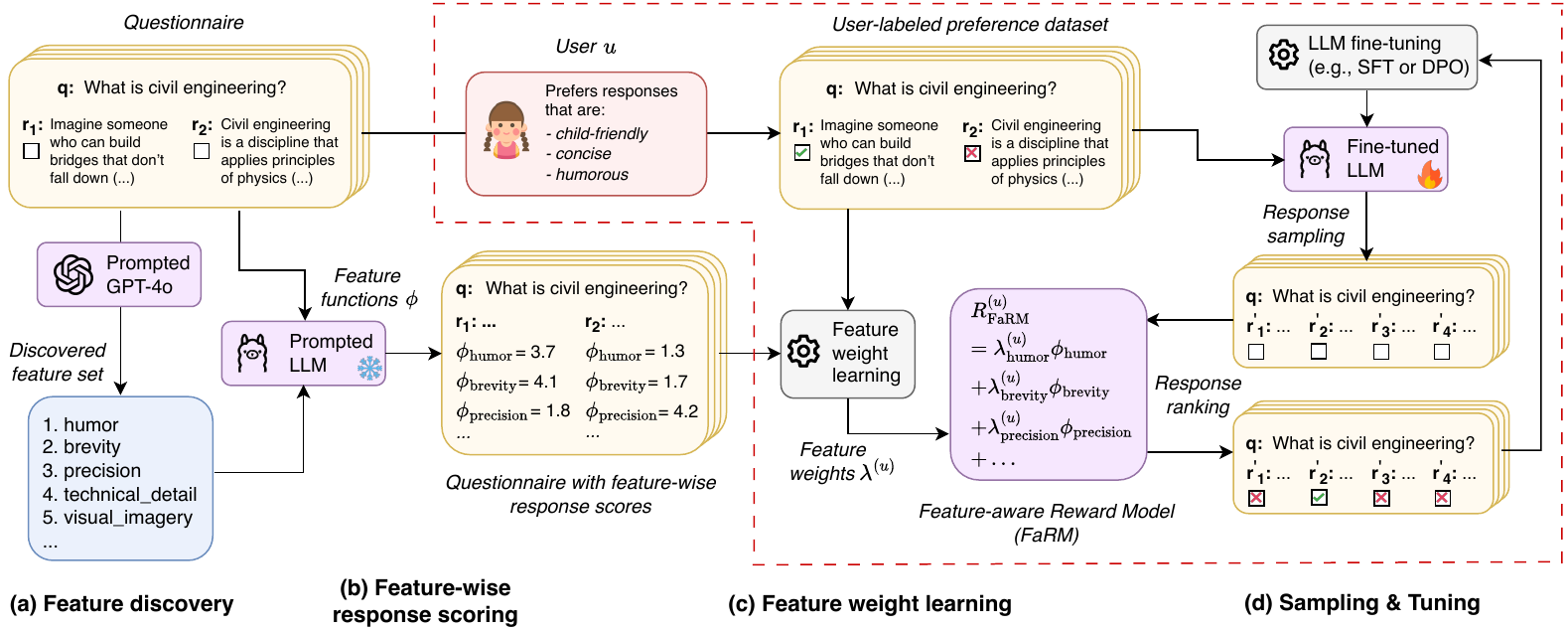} 
\caption{Overview of the proposed FaST approach. The red-dashed box highlights the user-specific steps.}
\label{fig:diagram}       
\end{figure*}

Our Feature-aware Sampling and Tuning (FaST) approach consists of two main stages: learning a highly parameter-efficient Feature-aware Reward Model (FaRM) from user-specific preference data, and training a generation model through a sampling-and-tuning procedure. An overview of FaST is shown in Fig.~\ref{fig:diagram}. The construction of our reward model is further decomposed into three steps: (a) discovering a set of interpretable features from the user-agnostic questionnaire, (b) scoring responses along these feature dimensions via prompted LLM-based feature functions, and (c) learning user-specific weights over the features using the user's expressed preferences within the questionnaire, resulting in a personalized Feature-aware Reward Model as a weighted average of feature functions. The subsequent sampling and tuning phase (d) proceeds by iteratively (i) sampling candidate responses, (ii) ranking them with FaRM, and (iii) fine-tuning the generation model on the ranked samples using, e.g., Supervised Fine-Tuning (SFT) or Direct Preference Optimization (DPO)~\citep{rafailov2023dpo}. The different steps of FaST are detailed in the following subsections.

\subsection{Feature Discovery}
\label{sec:feature-discovery}

The first step for building our Feature-aware Reward Model is to identify a set of relevant features from the available questionnaire composed of contexts and their associated possible responses. While in certain cases one may assume the set of features to be already known,\footnote{Existing reward models based on features or attributes~\cite{jang2023personalized,go2024compositional,wang2024arithmetic,Zhou2024,kim2025driftdecodingtimepersonalizedalignments} make this assumption and do not provide a methodology to identify relevant features.} this however does not hold true in general. One could possibly hire an expert in the specific domain of the dataset and ask them to identify some salient dimensions to be used as features, but this process may be costly and cumbersome. %

Instead, we propose a simple method for discovering data-informed features by including the entire questionnaire into the prompt of an LLM (namely, GPT-4o in the context of our experiments) and requesting it to provide a set of $F$ global features that characterize the specificities of the different responses. Note that this strategy can be carried out in the context of \ppalli{} because of the small size of the questionnaire (fewer than 100 questions) and the long-context abilities of modern LLMs. The prompt we crafted (detailed in App.~\ref{app:prompts}, Table~\ref{tab:prompt-discovery}) was designed to be \textit{generic}, so that it can be used for a questionnaire or a preference dataset from any domain or field of application (e.g., for role-play or question answering). Moreover, we emphasize that the features discovered by the prompted LLM are user-agnostic: the data provided in the prompt only contains the contexts and possible responses~-- no user-specific \textit{chosen} response label is used.

\subsection{Feature Function Definition}
\label{sec:response-scoring}

Once the relevant features of the dataset have been identified, we seek to annotate every response in the questionnaire with the value of each corresponding feature~-- which will further be used to learn the Feature-aware Reward Model from user preference data. For example, if one feature pertains to the humorous quality of a response, we wish to provide a score between 1 and 5 to express to what extent the response is funny (i.e., 1 means the response is completely serious and formal and 5 means it uses a highly humorous or playful tone). For that purpose, we follow a methodology similar to the one adopted in the Compositional Preference Model~\citep{go2024compositional}: we score a response with respect to a given feature by using a small\footnote{While a small size is not required per-se, the fact that the LLM-based scoring is required during generation model fine-tuning justifies some efficiency constraint on this operation.} LLM with a feature-specific prompt. In the remainder, we will refer to such prompted LLM as a \textit{feature function} and denote it as $\phi_f$ for a given feature $f$. The feature-specific fields of the prompt used in the feature function correspond to the overall description of the feature (e.g., ``How much humor or playfulness is present in the explanation?'') along with a natural language interpretation of the minimum and maximum scores (e.g., ``being completely serious and formal'' and ``using a highly humorous or playful tone''), which have all been automatically generated beforehand in the feature discovery step. In addition, the prompt also contains the context and the response to be scored.

While \citet{go2024compositional} obtain the feature-wise response score by extracting the numeric score from the text generated by the feature function, we argue that this method presents two drawbacks. First, this can be computationally inefficient if the feature function is verbose, as it requires to wait for the generation to complete before parsing the numeric score. Second, it does not account for the uncertainty that the LLM may have on the generated score. There may be cases where the LLM would have a relatively close probability for the tokens corresponding to two scores (e.g., 0.4 for score 3 and 0.55 for score 4) and retaining only the score with the highest likelihood leads to overlooking this uncertainty. Instead, the feature functions we use in FaST are designed such that (1) the numeric score is the next expected token after the prompt, and (2) score uncertainty is accounted for. The first design choice enables efficient scoring by requiring only a single forward pass through the LLM, after which the score token can be directly decoded. To address uncertainty, we compute a probability-weighted average over all score tokens -- rather than relying solely on the most probable one -- following a method similar to the one used for LLM-based evaluation in G-Eval~\cite{liu2023g-eval}. The prompt template used for the feature functions is detailed in App.~\ref{app:prompts}, Table~\ref{tab:prompt-scoring}.

\subsection{Feature-aware Reward Model}
\label{sec:farm}

In this section, we detail how we obtain FaRM by learning a personalized vector quantifying the weight of each feature to represent the individual preference of the user. Based on the preferred responses chosen by the user for each context of the questionnaire, we frame the weight learning problem as conditional log-likelihood estimation. Formally, let $\mathcal{D}^{(u)} = \{(q_i, r_{i,1}, \ldots, r_{i,K}, r^{(u)}_i)\}_{i=1}^D$ denote the preference dataset for user $u$ consisting of a set of $D$ tuples with a context $q_i$, the associated $K$ possible responses $r_{i,1}, \ldots, r_{i,K}$, and the user-chosen response $r^{(u)}_i \in \{r_{i,1}, \ldots, r_{i,K}\}$. Further denoting as $\lambda^{(u)} \in \mathbb{R}^F$ the feature weight vector to be learned for user $u$ with $F$ the number of discovered features, and $p(r \mid q; \lambda^{(u)})$ the probability that user $u$ prefers response $r$ among the options $r_{1}, \ldots, r_{K}$ available for context $q$, we formulate the following optimization problem:
\begin{equation}
\label{eq:opt}
    \max_{\lambda^{(u)}} \sum_{i=1}^D \log p(r^{(u)}_i \mid q_i; \lambda^{(u)})
\end{equation}
Intuitively, this problem states that we look for the $\lambda^{(u)}$ that maximizes the conditional likelihood of picking the user-chosen response $r^{(u)}_i$ among all options $r_{i,1}, \ldots, r_{i,K}$ available for context $q_i$. This objective can also be seen as a natural extension of the commonly used Bradley-Terry choice model~\citep{bradley1952rank} to the case where one response has to be picked among $K$ options instead of 
two~-- which is referred to as the McFadden choice model in the literature~\citep{mcfadden1974conditional}.

The conditional probability $p(r \mid q; \lambda^{(u)})$ is expressed in the following form:
\begin{equation}
\label{eq:conditional-prob}
    p(r \mid q; \lambda^{(u)}) = \frac{\exp\left(R_{\text{FaRM}}^{(u)}(q, r)\right)}{\sum_{k=1}^K \exp\left(R_{\text{FaRM}}^{(u)}(q, r_k)\right)}
\end{equation}
where the term $R_{\text{FaRM}}^{(u)}(q, r) = \phi(q, r)^T\lambda^{(u)} = \sum_{i=1}^F  \lambda^{(u)}_{f_i} \, \phi_{f_i}(q, r)$ defines the Feature-aware Reward Model associated to user $u$. Here, $\phi(q, r) = [\phi_{f_1}(q, r), \ldots, \phi_{f_F}(q, r)] \in \mathbb{R}^F$ is the vector of feature-wise scores given by the feature functions for every feature $f$ applied to context $q$ and response $r$. %
FaRM is thus a linear combination of feature-specific frozen reward models with personalized feature weights~-- these $F$ weights constituting the only learned parameters of FaRM, yielding a highly parameter-efficient approach. This form is inspired by multi-objective reward models~\citep{jang2023personalized,wang2024arithmetic} and specifically by the Compositional Preference Model~\citep{go2024compositional}. Plugging Eq.~\ref{eq:conditional-prob} into the objective function~\eqref{eq:opt} leads to a convex optimization problem (see App.~\ref{app:convexity} for a proof of its convexity) from which we learn the vector $\lambda^{(u)}$ via gradient descent.

\subsection{Generation Model Fine-tuning}
\label{sec:finetuning}

After learning the Feature-aware Reward Model for the user of interest, we use it to fine-tune an LLM and enable personalized generation for this user. The fine-tuning method we adopt here is referred to as \textit{sampling and tuning}, which can be seen as a generic framework regrouping several iterative fine-tuning approaches such as Rejection sampling Fine-tuning (RFT)~\citep{dong2023raft,touvron2023llama2} and Online DPO~\citep{guo2024online-dpo}.

Given a base LLM we wish to fine-tune and a reward model (FaRM, in our case), sampling and tuning iterates over the following steps: (i) for each context in the preference training set, $S$ candidate responses are drawn from the LLM, (ii) candidate responses are scored and ranked using the reward model, and (iii) the LLM is updated by fine-tuning it based on the ranked candidate responses. Using SFT for the fine-tuning in step (iii) leads to RFT, and using DPO instead results in Online-DPO. In our implementation, we add to the list of candidate responses the response that obtained the highest reward in the previous iteration (or the user-chosen response from the initial preference dataset for the first iteration), in order to enforce a monotonous progression in the best reward obtained over successive iterations.

We have adopted this fine-tuning approach instead of the commonly used Proximal Policy Optimization (PPO)~\citep{schulman2017ppo} for several reasons. First, RFT (and by extension, sampling and tuning) was shown in various works~\citep{touvron2023llama2,xiong2025minimalistapproachllmreasoning} to be competitive against PPO while being more stable and having much fewer hyperparameters to tune. Secondly, unlike PPO, sampling and tuning only requires a ranking of the candidate responses rather than the raw reward values~-- whose range may vary across users depending on the obtained $\lambda^{(u)}$, in turn affecting the stability of the fine-tuning algorithm. Finally, sampling and tuning is more parameter-efficient than PPO~-- which requires the learning of value function networks~-- making the former more suitable for our very low-data regime.

\section{Experiments}

\subsection{Datasets}
\label{sec:datasets}

To benchmark preference alignment approaches on the \ppalli{} problem for questionnaire-based LLM personalization, we introduce and release two new preference datasets: DnD and ELIP.
In the remainder of this subsection, we first explain the motivation behind creating these datasets before providing a brief description of DnD and ELIP. We further detail their construction in App.~\ref{app:dataset}.

\begin{table*}[t]
\centering
\scalebox{0.9}{
\begin{tabular}{lccc}
\toprule
\textbf{Dataset} & \textbf{\begin{tabular}[c]{@{}c@{}}User-agnostic\\ contexts \& responses\end{tabular}} & \textbf{\begin{tabular}[c]{@{}c@{}}User-specific\\ preference annotations\end{tabular}} & \textbf{\begin{tabular}[c]{@{}c@{}}Interpretable\\ user profiles\end{tabular}} \\ \midrule
Personalized Soups~\citep{jang2023personalized} & {\color{ForestGreen} \cmark} & {\color{BrickRed} \xmark} & {\color{ForestGreen} \cmark} \\
PERSONA~\citep{castricato2024persona} & {\color{BrickRed} \xmark} & {\color{ForestGreen} \cmark} & {\color{ForestGreen} \cmark} \\
PRISM~\citep{kirk2024prism} & {\color{BrickRed} \xmark} & {\color{ForestGreen} \cmark} & {\color{ForestGreen} \cmark} \\
Perspective~\citep{kim2025driftdecodingtimepersonalizedalignments} & {\color{BrickRed} \xmark} & {\color{ForestGreen} \cmark} & {\color{ForestGreen} \cmark} \\
PersonalLLM~\citep{zollo2024personalllm} & {\color{ForestGreen} \cmark} & {\color{ForestGreen} \cmark} & {\color{BrickRed} \xmark} \\ \midrule
DnD (ours) & {\color{ForestGreen} \cmark} & {\color{ForestGreen} \cmark} & {\color{ForestGreen} \cmark} \\
ELIP (ours) & {\color{ForestGreen} \cmark} & {\color{ForestGreen} \cmark} & {\color{ForestGreen} \cmark} \\ \bottomrule
\end{tabular}
}
\caption{Comparison of the characteristics of existing personalized preference datasets and our proposed datasets, DnD and ELIP.}
\label{tab:datasets}
\end{table*}

\paragraph{Limitations of existing datasets.} 

As mentioned in Section~\ref{sec:problem}, existing personalized preference datasets~-- namely, Personalized Soups~\citep{jang2023personalized}, PERSONA~\cite{castricato2024persona}, PRISM~\citep{kirk2024prism}, and Perspective~\citep{kim2025driftdecodingtimepersonalizedalignments}~-- are not suitable for the questionnaire-based setting proposed in \ppalli{}. These datasets either rely on user-specific contexts, user-specific responses, or both, and thus do not employ a shared questionnaire format to elicit user-specific preferences. The only dataset that follows a questionnaire-based approach similar to ours is PersonalLLM~\citep{zollo2024personalllm}. However, its users are simulated through linear combinations of open-source reward models, resulting in preferences that are highly artificial and do not lend themselves to a transparent interpretation of the generation results as the users are entirely latent. Due to these limitations, we created two small-scale, high-quality datasets~-- DnD and ELIP~-- that align more closely with our personalized, questionnaire-driven preference alignment task. The comparison of existing personalized preference datasets with our proposed datasets is summarized in Table~\ref{tab:datasets}, along three dimensions: (i) whether the dataset relies on a user-agnostic questionnaire with shared contexts and responses, (ii) whether preference annotations are provided for each user, and (iii) whether user profiles are interpretable and associated with natural language characteristics.

\paragraph{DnD.} 

Our DnD (``Dungeons and Dragons'') dataset focuses on a role-playing task where the goal is to simulate the actions of a character from a fantasy universe when facing in-game situations. Our dataset is composed of 10 characters with diverse characteristics (in terms of race, class, moral alignment, etc.) and 129 in-game situations associated with 3 possible actions each. For each of the 1,290 (character, situation) pairs, we provide an annotation of the action preferred by the character among the possible options for this situation. The generation of the character descriptions, situations, actions and the preferred action annotation have all been done using GPT-4o. 

\paragraph{ELIP.} 

The second dataset, ELIP (``Explain Like I Prefer''), defines a conversational assistant scenario where the responses to open-ended 
questions should be tailored to the user preferences. We used a set of 100 human-curated questions from the ELI5 (``Explain Like I'm 5'') dataset~\citep{fan2019eli5}~-- selected for their open-ended nature while maintaining some topic diversity. For each question, we generated 4 diverse possible responses using GPT-4o. The user preferences were inspired by the 3 dimensions introduced in the Personalized Soups dataset~\citep{jang2023personalized}: expertise (\textit{child-friendly} vs \textit{expert-level}), informativeness level (\textit{concise and to-the-point} vs \textit{detailed}), and tone (\textit{friendly and humorous} vs \textit{impersonal}). Considering all possible combinations of these dimensions yielded 8 different users. Preferred response annotation for each of the 800 (user, question) pairs was performed with GPT-4o as well.

\subsection{Preferred Response Prediction}

\paragraph{Experimental setup.}

We first validate the proposed reward model, FaRM, by measuring its ability to predict the response preferred by a user on unseen contexts. This task can therefore be seen as classification, where the user's feature weight vector in FaRM is learned from the train split and the resulting FaRM is used to predict the responses preferred by this user on the validation and test contexts. For both DnD and ELIP, we use a 50\%/25\%/25\% split for train/validation/test and report the average results over 5 random splits. Performance is measured by the accuracy of the \textit{chosen} response being predicted. 

\paragraph{Baselines.}

We compared \textbf{FaRM} against the following baselines: a \textbf{Random} classifier; a \textbf{Manyshot} in-context learning classifier~\citep{agarwal2024manyshot} which predicts a response to a context based on a prompt containing the full training set of the user preference data; a chain-of-thought version of the latter, \textbf{Manyshot-CoT}, which generates a description of the user before making the prediction; a traditional reward model (\textbf{RM}) obtained by fully fine-tuning an LLM stacked with a scalar head, as well as a parameter-efficient variant based on low-rank adapters~\citep{hu2022lora}, \textbf{RM-LoRA}; the Compositional Preference Model (\textbf{CPM}) proposed by~\citet{go2024compositional}, which is applied to the features discovered by FaRM. We tested different backbone models for the frozen feature functions used in FaRM and CPM, for the base LLM fine-tuned within RM, and for the prompted LLM used in the manyshot baselines: LLaMA-3.2-3B-Instruct and Phi-4-Mini-Instruct.\footnote{Available at \url{https://huggingface.co/meta-llama/Llama-3.2-3B-Instruct} and \url{https://huggingface.co/microsoft/Phi-4-mini-instruct}, respectively.} FaRM was based on $F = 20$ features (shown in App.~\ref{app:features}, Tables~\ref{tab:dnd-features} and~\ref{tab:elip-features}). Other hyperparameters for FaRM and RM are detailed in App.~\ref{app:hyperparameters}.

\begin{table*}[t]
\centering
\scalebox{0.85}{
\begin{tabular}{@{}lcccccccccccc@{}}
\toprule
\multirow{4}{*}{\textbf{Preference model}} & \multicolumn{6}{c}{\textbf{Acc. (\%) on DnD}} & \multicolumn{6}{c}{\textbf{Acc. (\%) on ELIP}} \\ \cmidrule(l{4pt}r{4pt}){2-7} \cmidrule(l{4pt}r{4pt}){8-13}
 & \multicolumn{3}{c}{\textbf{LLaMA-3.2-3B-It}} & \multicolumn{3}{c}{\textbf{Phi-4-Mini-It}} & \multicolumn{3}{c}{\textbf{LLaMA-3.2-3B-It}} & \multicolumn{3}{c}{\textbf{Phi-4-Mini-It}} \\ \cmidrule(l{4pt}r{4pt}){2-4} \cmidrule(l{4pt}r{4pt}){5-7} \cmidrule(l{4pt}r{4pt}){8-10} \cmidrule(l{4pt}r{4pt}){11-13}
 & \textbf{train} & \textbf{val} & \textbf{test} & \textbf{train} & \textbf{val} & \textbf{test} & \textbf{train} & \textbf{val} & \textbf{test} & \textbf{train} & \textbf{val} & \textbf{test} \\ \midrule
Random & 33.3 & 33.3 & 33.3 & 33.3 & 33.3 & 33.3 & 25.0 & 25.0 & 25.0 & 25.0 & 25.0 & 25.0 \\
Manyshot & 52.5 & 41.6 & 40.9 & {\ul 83.3} & 45.1 & 45.5 & 41.9 & 25.0 & 26.8 & 50.3 & 25.4 & 30.0 \\
Manyshot-CoT & 42.8 & 37.7 & 38.0 & 52.0 & 40.2 & 43.7 & 48.6 & 31.5 & 33.0 & 66.6 & 32.9 & 37.5 \\
RM & \textbf{97.9} & {\ul 60.6} & {\ul 62.6} & \textbf{96.0} & 59.6 & 63.9 & \textbf{99.8} & \textbf{73.3} & {\ul 70.9} & \textbf{99.6} & 72.8 & 72.0 \\
RM-LoRA & {\ul 78.7} & 57.0 & 58.8 & 73.4 & 58.3 & 60.2 & {\ul 86.1} & 67.5 & \textbf{71.0} & {\ul 81.7} & 72.6 & 71.4 \\
CPM & 68.5 & 58.5 & 58.5 & 72.7 & {\ul 63.8} & {\ul 66.5} & 67.5 & 65.4 & 60.6 & 77.3 & {\ul 74.1} & {\ul 72.8} \\
FaRM (ours) & 71.7 & \textbf{65.5} & \textbf{63.9} & 75.3 & \textbf{66.6} & \textbf{69.4} & 75.9 & {\ul 71.1} & \textbf{71.0} & 80.6 & \textbf{76.1} & \textbf{75.3} \\
\bottomrule
\end{tabular}
}
\caption{Preferred response prediction results on DnD and ELIP, in terms of accuracy (higher is better). The best results are shown in bold, and the second-best ones are underlined.}
\label{tab:rp-main}
\end{table*}

\begin{figure}[t]
\centering
\includegraphics[width = .37\textwidth]{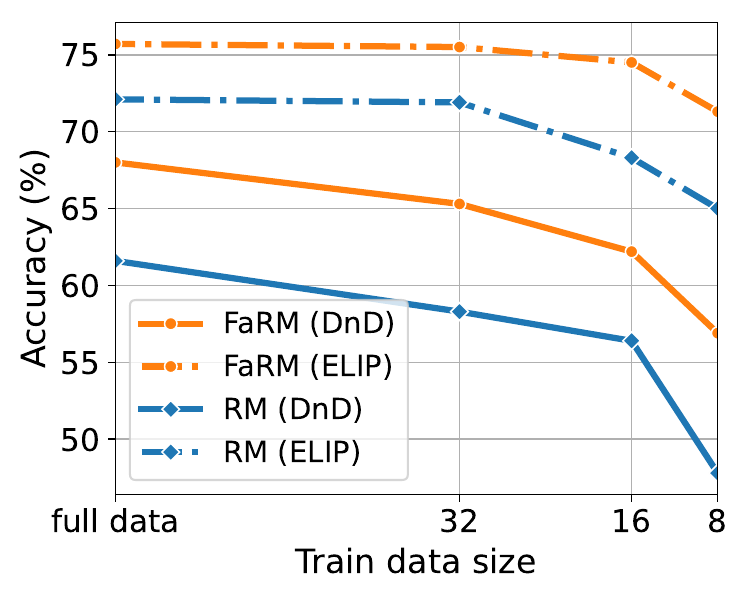}
\caption{Comparison of the preferred response prediction accuracy (higher is better) obtained by RM and FaRM on DnD and ELIP, across different train set sizes. Full training data corresponds to 64 contexts for DnD and 50 contexts for ELIP.}
\label{fig:rp-vary-size} 
\end{figure}

\paragraph{Results.}

Table~\ref{tab:rp-main} presents the main results for the preferred response prediction task. The in-context learning baselines, Manyshot and Manyshot-CoT, performed poorly, with an accuracy near random chance on both the validation and test sets. We attribute this to the unnatural task format for an LLM: selecting a response by outputting a numeric ID is misaligned with the training paradigm of LLMs, which are optimized for generating natural language rather than discrete classification. In contrast, RM, RM-LoRA, CPM and FaRM performed substantially better. Notably, FaRM achieved highly competitive results despite having a minimal number of learned parameters (equal to the number of features, $F$). This compactness likely contributed to its greatly reduced overfitting compared to RM, particularly given the small training set sizes (64 for DnD and 50 for ELIP). While RM-LoRA optimizes significantly fewer parameters than RM, which may reduce the risk of overfitting, its performance still remains inferior to RM on the validation and test sets in most cases. Overall, FaRM achieved the highest validation and test accuracy, outperforming RM, RM-LoRA and CPM by a good margin. 

\paragraph{Efficiency and robustness.}

FaRM's competitive effectiveness is also combined with a greatly improved efficiency over RM~-- for instance, learning FaRM's feature weights for the 8 users of ELIP took around 7 seconds on a single CPU machine, while fine-tuning RM on the same dataset took around 50 minutes with a A100 GPU. Additionally, we studied the performance of RM and FaRM across varying training data sizes~-- using either the full dataset or subsets containing 32, 16, or 8 contexts. Figure~\ref{fig:rp-vary-size} presents the corresponding accuracy results averaged over the validation and test set, focusing on the best-performing configurations identified for RM and FaRM in Table~\ref{tab:rp-main}: RM w/ LLaMA-3.2-3B-Instruct and FaRM w/ Phi-4-Mini-Instruct. The figure shows that FaRM consistently outperforms RM across all data sizes. Moreover, FaRM demonstrates greater robustness to reduced training size on the ELIP dataset, maintaining accuracy even with as few as 16 instances. In addition to these results, we also provide a detailed analysis in App.~\ref{app:farm} on different FaRM variants and ablations to confirm the importance of its components.

\subsection{Personalized Generation}
\label{sec:generation-exp}

\paragraph{Experimental setup.} The personalized generation task involves generating user-tailored responses for unseen contexts, guided solely by user preferences derived from chosen responses on a shared questionnaire. We emphasize that no explicit user information~-- such as descriptions or profiles~-- is provided; only user preference data is available. Performance is measured using two metrics given by an LLM-judge: a 5-point personalization score inspired by rubric-based LLM evaluation~\citep{kim2023prometheus,thonet2025elitr-bench}, and a winrate computed from pairwise comparisons of model outputs~\citep{liusie2024}. To limit the number of comparisons, each method is compared only against a weak baseline (Zeroshot) and an oracle (Oracle-chosen), both described below.

\paragraph{Baselines.}

Our experiments compare the following approaches: the non-personalized \textbf{Zeroshot} LLM; in-context learning (ICL) approaches including retrieval-augmented generation (\textbf{RAG}) from similar training contexts, \textbf{Manyshot}~\citep{agarwal2024manyshot} and \textbf{Manyshot-CoT}; fine-tuning with no explicit reward model: \textbf{SFT} and \textbf{DPO}~\citep{rafailov2023dpo}; and approaches based on a traditional reward model (\textbf{RM}) combined with \textbf{Best-of-N}, \textbf{PPO}~\citep{schulman2017ppo}, \textbf{Online-DPO}~\citep{guo2024online-dpo}, and \textbf{RFT}~\citep{dong2023raft}. The proposed \textbf{FaST} is also instantiated with \textbf{Best-of-N}, \textbf{Online-DPO}, and \textbf{RFT}.\footnote{In our personalized generation experiments with FaST and RM-based approaches, we utilized the preference model which yielded the best validation accuracy in the preferred response prediction task (respectively, FaRM w/ Phi-4-Mini-It and RM w/ LLaMA-3.2-3B-It).} The LLM prompted or fine-tuned in all these approaches is LLaMA-3.2-3B-Instruct. We include as well two oracle methods based on GPT-4o which showcase the performance that can be achieved when having access to the explicit user description: \textbf{Oracle-chosen} simply returns the response labeled as \textit{chosen} (among the available options) for the user during the construction of the dataset, while \textbf{Oracle-gen} is prompted with the user description to explicitly generate a tailored response. More details on evaluation, baselines, and hyperparameters can be found in App.~\ref{app:exp-setup}.

\paragraph{Results.}

\begin{figure*}[t]
\centering
\begin{minipage}[b]{0.69\textwidth}
\centering
\scalebox{0.82}{
\begin{tabular}{llcccccc}
\toprule
\multicolumn{2}{c}{\multirow{2}{*}{\textbf{Approach}}}  & \multicolumn{3}{c}{\begin{tabular}{@{}c@{}}\textbf{Score improv.}\\\textbf{over Zeroshot}\end{tabular}} & \multicolumn{3}{c}{\begin{tabular}{@{}c@{}}\textbf{Winrate (\%) vs}\\\textbf{Zeroshot; Oracle-chosen}\end{tabular}} \\
\cmidrule(l{4pt}r{4pt}){3-5} \cmidrule(l{4pt}r{4pt}){6-8}
 &  & \textbf{DnD} & \textbf{ELIP} & \textbf{Avg} & \textbf{DnD} & \textbf{ELIP} & \textbf{Avg} \\ \midrule
 \multirow{2}{*}{\textit{Oracle}} & Oracle-chosen & 0.37 & 0.59 & 0.48 & 74.8; 50.0 & 75.4; 50.0 & 75.1; 50.0 \\ 
 & Oracle-gen & 0.84 & 0.80 & 0.82 & 94.0; 81.8 & 95.8; 85.2 & 94.9; 83.5 \\
\arrayrulecolor{lightgray}\midrule\midrule
 \textit{Base} & Zeroshot & 0.00 & 0.00 & 0.00 & 50.0; 25.2 & 50.0; 24.6 & 50.0; 24.9 \\ \midrule
\multirow{3}{*}{\textit{ICL}}   & RAG & 0.17 & 0.30 & 0.24 & 65.9; 37.5 & 66.8; 33.2 & 66.3; 35.4 \\ 
& Manyshot & 0.24 & 0.30 & 0.27 & 73.6; 47.0 & 66.3; 32.5 & 70.0; 39.7 \\
 & Manyshot-CoT & 0.32 & 0.33 & 0.33 & 79.2; 54.4 & 70.3; 39.9 & 74.8; 47.2 \\ \midrule
\multirow{2}{*}{\textit{RM-free}} & SFT & 0.27 & 0.41 & 0.34 & 69.7; 42.7 & 68.3; 39.4 & 69.0; 41.1 \\
 & DPO & 0.32 & {\ul 0.54} & 0.43 & 74.9; 47.2 & \textbf{75.8}; \textbf{52.8} & 75.3; 50.0 \\ \midrule
\multirow{4}{*}{\textit{RM}} & w/ Best-of-N & 0.16 & 0.12 & 0.14 & 62.5; 33.2 & 53.2; 30.0 & 57.8; 31.6 \\
 & w/ PPO & 0.34 & 0.08 & 0.21 & 71.3; 40.4 & 46.7; 30.2 & 59.0; 35.3 \\ 
 & w/ Online-DPO & 0.35 & 0.33 & 0.34 & 69.3; 45.8 & 67.6; 43.9 & 68.4; 44.9 \\
 & w/ RFT & 0.37 & 0.49 & 0.43 & 74.0; 49.2 & {\ul 74.4}; 47.5 & 74.2; 48.4 \\ \midrule
\multirow{3}{*}{\begin{tabular}{@{}l@{}}\textit{FaST}\\\textit{(ours)}\end{tabular}} & w/ Best-of-N  & 0.25 & 0.19 & 0.22 & 72.1; 40.0 & 56.6; 29.2 & 64.4; 34.6 \\
 & w/ Online-DPO  & \textbf{0.48} & 0.42 & {\ul 0.45} & \textbf{82.1}; \textbf{62.3} & 73.4; {\ul 50.2} & \textbf{77.8}; \textbf{56.2} \\
 & w/ RFT & {\ul 0.46} & \textbf{0.57} & \textbf{0.51} & {\ul 80.5}; {\ul 58.9} & 73.8; 47.3 & {\ul 77.1}; {\ul 53.1} \\
\arrayrulecolor{black}\bottomrule
\end{tabular}
}
\captionof{table}{Personalized generation results on DnD and ELIP (higher is better). The reported personalization scores (left) correspond to the improvement over the Zeroshot baseline. The best results (excluding oracle approaches) are shown in bold, and the second-best ones are underlined.}
\label{tab:gen-main}
\end{minipage}%
\hfill
\begin{minipage}[b]{0.28\textwidth}
\centering
\includegraphics[width = .95\textwidth]{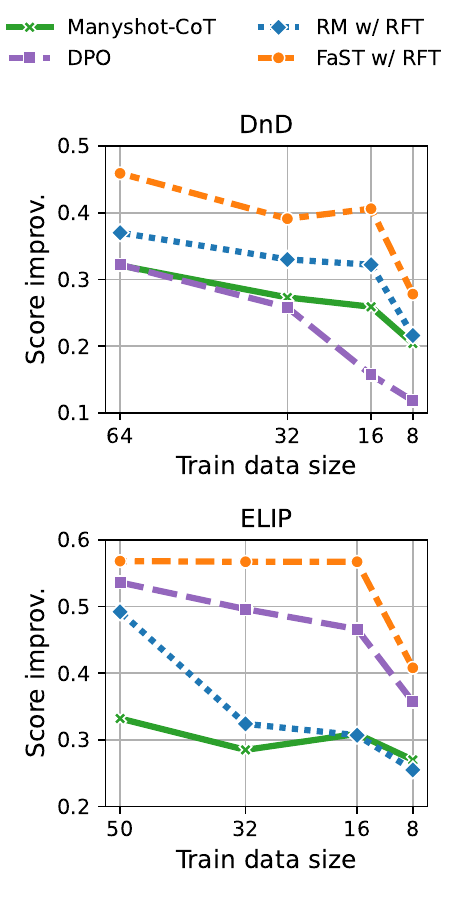}
\vspace{-0.2cm}
\caption{Personalized generation results across different train set sizes (higher is better).
}
\label{fig:gen-vary-size}
\end{minipage}
\end{figure*}

The main results for personalized generation are shown in Table~\ref{tab:gen-main}. On the left, we report the personalization scores as an offset with respect to the Zeroshot baseline (raw score results can be found in App.~\ref{app:gen-results}). Reported performance results from first averaging over users (10 for DnD and 8 for ELIP), then averaging over 3 train/validation/test splits, and finally averaging over the validation and test sets. We observe that overall, FaST w/ Online-DPO and RFT lead to the best results on average for DnD and ELIP and for the different metrics~-- although there is no clear winner between the Online-DPO and RFT variants. Interestingly, we note that FaST w/ RFT is even able to perform on par or better than Oracle-chosen for all metrics, despite having no access to privileged information about the user as the oracle method does. Among the ICL baselines, Manyshot-CoT performed the best, while for RM-free fine-tuning approaches, DPO obtained the best results. Notably, DPO obtained the overall best results on the winrate metrics for ELIP, but was still on average inferior to the best FaST variants. Comparing between RM-based and FaST variants, we observe that results are either in favor of FaST or comparable. Our different observations are supported by our complementary results (see App.~\ref{app:gen-results}), including the winrates across FaST/RM variant pairs (Table~\ref{tab:gen-wr-add}), the obtained Elo rankings (Table~\ref{tab:elo-rankings}), the results comparing full-model fine-tuning with LoRA-based fine-tuning (Table~\ref{tab:gen-scores-lora}), and the generation performance using the 1B-parameter version of the model (Table~\ref{tab:gen-scores-1b}).
We also examined how training set size affects generation performance. As shown in Fig.~\ref{fig:gen-vary-size}, FaST w/ RFT maintains strong performance on both DnD and ELIP with as few as 16 training instances, thus demonstrating greater robustness than competing methods.

\begin{table}[t]
\centering
\scalebox{0.8}{
\begin{tabular}{@{}p{1.5cm}p{3cm}p{3cm}@{}}
\toprule
\textbf{DnD} & & \\
\midrule
\multicolumn{3}{@{}p{9.8cm}@{}}{\textbf{Situation:} The lush valley you find yourself in is filled with vibrant, exotic flora and fauna. However, in the center of this paradise stands an imposing stone tower, covered in creeping vines. Smoke curls from the topmost window, and a faint smell of sulfur taints the air. The locals speak of a reclusive sorcerer who dwells within, guarding ancient secrets.} \\\addlinespace
\multicolumn{3}{@{}l@{}}{\textbf{Generated action:}} \\
\textit{Zeroshot:} & \multicolumn{2}{@{}p{8.1cm}@{}}{I draw my short sword and cautiously approach the tower's main entrance, searching for any signs of magical wards or hidden dangers.} \\
\textit{FaST: (Grog)} & \multicolumn{2}{@{}p{8.1cm}@{}}{I draw my sword and charge towards the tower's entrance, ready to confront the sorcerer himself.} \\
\midrule
\midrule
\textbf{ELIP} & & \\
\midrule
\multicolumn{3}{@{}p{9.8cm}@{}}{\textbf{Question:} What is civil engineering?} \\\addlinespace
\multicolumn{3}{@{}l@{}}{\textbf{Generated response:}} \\
\textit{Zeroshot:} & \multicolumn{2}{@{}p{8.1cm}@{}}{Civil engineering is the application of scientific and mathematical principles to design, build, and maintain infrastructure, such as roads, bridges, buildings, and water treatment facilities, with a focus on ensuring public safety and well-being.} \\
\textit{FaST: (AAA)} & \multicolumn{2}{@{}p{8.1cm}@{}}{Imagine planning and building the superhighways, bridges, and skyscrapers of our world—civil engineering is the team of experts who make it all happen!} \\
\bottomrule
\end{tabular}
}
\caption{Example generations from (non-personalized) Zeroshot, and FaST personalized for $u$ = ``Grog'' (on DnD) and ``AAA'' (on ELIP). The character Grog is a \textit{chaotic-neutral half-orc barbarian} who favors aggressive and direct actions when tackling DnD situations, and user AAA corresponds to a profile that prefers \textit{child-friendly}, \textit{concise} and \textit{humorous} responses in their interactions with a conversational assistant.}
\label{tab:samples}
\end{table}

\paragraph{Generated examples.}

Table~\ref{tab:samples} presents example responses generated by Zeroshot and FaST w/ RFT. The responses shown for FaST correspond to the character ``Grog'' on DnD (a \textit{chaotic-neutral half-orc barbarian} characterized by his aggressive behavior) and the user identified as ``AAA'' on ELIP (who prefers \textit{child-friendly}, \textit{concise} and \textit{humorous} responses). These examples demonstrate that FaST produces responses more aligned with each profile's intuitive traits, whereas the Zeroshot responses appear noticeably less tailored. Additional responses can be found in App.~\ref{app:samples}, Fig.~\ref{tab:samples-extended}.

\paragraph{Personalization for under-represented users.}

For certain users, we found user-chosen responses to be misaligned~-- shown by lower personalization scores for Oracle-chosen on these users. 
For instance, the character ``Grog'' from DnD received a score of 2.868 out of 5~-- well below the average of 3.456 obtained by Oracle-chosen across all characters~-- suggesting a mismatch between his preferences and the available options which Oracle-chosen picked from. Grog's direct and aggressive traits likely diverge from the socially calibrated behaviors typical of LLMs like GPT-4o, which generated the Oracle-chosen responses. In contrast, FaST notably improved Grog's personalization score by learning his latent preferences from generally misaligned choices. FaST w/ RFT achieved 3.298, approaching the method's overall average of 3.552, with a 70.7\% winrate over Oracle-chosen (vs 58.9\% average). We attribute this to FaST's use of high-level features, enabling it to model diverse user preferences~-- even those that deviate from the norm. This highlights FaST's potential %
to bridge the gap between satisfying mainstream preferences and better serving under-represented users~-- an essential property for ensuring fairness in personalized generation from LLMs.

\section{Related Work}

\paragraph{Personalized preference alignment.}

Reinforcement Learning from Human Feedback~\citep{Ouyang2022} and DPO~\citep{rafailov2023dpo} are standard methods for aligning LLMs with general human preferences, but the challenges of adapting them to individual users~-- i.e., learning user-specific policies or reward models~--  have motivated recent research in personalized alignment~\citep{Guan2025}. Some approaches, like Personalized Soups~\citep{jang2023personalized} and P-DPO~\citep{Li2024}, learn user-specific models via low-dimensional weights or embeddings. Other recent concurrent works, like PAD~\citep{chen2025pad}, MoPE~\citep{Zhou2024}, AMULET~\citep{zhang2025amulet}, and Drift~\citep{kim2025driftdecodingtimepersonalizedalignments}, %
personalize at inference-time by reweighting token logits. In contrast, FSPO~\citep{singh2025fspofewshotpreferenceoptimization} studies a few-shot approach which treats preference learning as meta-learning. Similarly to us, using high-level features or attributes is adopted by several personalized alignment methods~\citep{jang2023personalized,Zhou2024,kim2025driftdecodingtimepersonalizedalignments}. However, these assume that the relevant features are already identified and known~-- whereas FaST discovers them directly from the questionnaire data.

\paragraph{Preference alignment from limited data.}

Recent work on aligning language models from limited data explores diverse methodologies to efficiently utilize minimal preference annotations. Approaches like ALMA~\citep{yasunaga2024alma} and Spread~\citep{kim2025spread} synthesize preferences from small seed sets, while others such as DELIFT~\citep{agarwal2025delift} and~\citet{deng2025less} prioritize informative annotations. Though data-efficient methods still rely on $\sim$1000 annotations~\citep{jin2023data}, the proposed FaST achieves robust alignment with fewer than 100.

\section{Conclusion}

We introduced the problem of Personalized Preference Alignment with Limited Data, a practical scenario where only a small set of annotations from user-shared questionnaires is available for tailoring LLM outputs to individual users. To support research in this setting, we proposed two new datasets~-- DnD and ELIP. We further presented FaST, a highly parameter-efficient approach that leverages automatically discovered features to enable effective alignment from minimal user data. Extensive experiments on both datasets show that FaST achieves competitive or superior personalization compared to strong baselines, while being robust under data constraints. Lastly, we underlined the potential of this approach to better serve under-represented users, offering a path toward fairer and more inclusive LLM personalization.

\section*{Limitations}

While the proposed FaST approach demonstrates strong personalization performance in the studied \ppalli{} setting, it also comes with several limitations that warrant discussion. 

First, our sampling-and-tuning procedure relies on the ability of the base LLM to generate a sufficiently diverse set of candidate responses. This diversity is critical for identifying outputs that yield increasing FaRM rewards in the iterative process, in order to progressively improve the personalization capabilities of the learned policy. Although adjusting the sampling temperature can help encourage sample diversity, it may be insufficient when the desired outputs fall outside the distribution supported by the base model. Importantly, however, this challenge is not unique to FaST; it is shared by all sampling-based fine-tuning methods, including those based on PPO, RFT, and Online-DPO. As a potential solution, we plan to explore in future work a sampling scheme which involves prompting the policy to generate all candidate responses jointly as a list, using a prompt explicitly crafted for ensuring diversity across the candidates.

Second, our evaluation of personalized generation relies heavily on LLM-judges. While such evaluation method inherently introduces some inconsistencies and variability, we try and mitigate these limitations by employing two complementary metrics: a pointwise personalization score, and the pairwise win-rate comparisons against a weak baseline (Zeroshot) and a strong reference (Oracle-chosen). We also derive an Elo ranking from a broader set of pairwise matchups. These different metrics generally yield consistent trends, lending confidence to our reported findings. %

Finally, the user profiles in the ELIP dataset~-- adapted from~\citet{jang2023personalized}~-- rely on simple preference dimensions %
to enable controlled experimentation. While this facilitates benchmarking and result interpretation, it may not fully capture the richness or variability of real user preferences. As a direction for future work, we will consider conducting user studies in which participants provide preference annotations on ELIP training and validation contexts, and rate the personalized generations on held-out test contexts. This would offer a more realistic assessment of personalization effectiveness and user satisfaction.

\section*{Ethical statement}

Fine-grained personalization of a conversational assistant based on just a few dozen user preference annotations is a powerful capability. As such, it must be handled with great care due to the sensitive nature of personal information. If misused, it could manipulate user opinions~-- whether influencing political views during an election or promoting deceptive products in fraudulent commercial schemes~-- and reinforce so-called filter bubbles~\citep{Pariser2011}.

However, by relying on high-level, interpretable features, the proposed FaST approach naturally lends itself to mitigating some of these risks. Indeed, this design choice promotes greater transparency and user agency, as it enables explicit disclosure of the personalization dimensions to users. Users can~-- and \textit{should}~-- be informed about which features are influencing the assistant's behavior and could potentially opt out of specific features they are uncomfortable with. This feature-based methodology enhances accountability and sets our approach apart from the standard RLHF framework~\citep{Ouyang2022} which relies on black-box reward models that remain entirely opaque in their decision making. While we cannot prevent all potential misuses, we believe that designing for transparency through explicit features represents a responsible contribution that better enables ethical implementations.

Furthermore, personalization opens valuable opportunities to better serve marginalized and under-represented communities, who are often overlooked by models optimized for majority preferences. In particular, our observations highlight that personalization with FaST can improve alignment with the needs of such users, thus fostering more equitable and inclusive interactions.

\section*{Acknowledgments}

We thank Marcely Zanon Boito for her valuable suggestions on the design of the DnD dataset. We are also grateful to Florent Perronnin and Laurent Besacier for their insightful discussions and feedback on this work.

{
    \small
    \bibliographystyle{ieeenat_fullname}
    \bibliography{main}
}

\clearpage
\appendix

\section{Convexity of the Feature Weight Learning Objective}
\label{app:convexity}

Following \eqref{eq:opt}, we want to prove that the following sum is a convex function of $\lambda$:
\begin{equation}
\label{eq:app-convex-sum-over-dataset}
 \sum_{i=1}^D - \log p(r_i \mid q_i; \lambda),
\end{equation}
where
\begin{equation*}
    p(r \mid q; \lambda) = \frac{e^{\phi(q, r)^T\lambda}}{\sum_{k=1}^K e^{\phi(q, r_k)^T\lambda}},
\end{equation*}
and where the $r_k$ are the different responses in competition for the given context $q = q_i$, with $r = r_i$ the preferred response.
Defining $z_k = \phi(q, r_k)$, we have $-\log p(r \mid q; \lambda) = - z_k^T\lambda + \log  \sum_{k=1}^K e^{z_k^T\lambda}$. Because the first term is a linear function of $\lambda$, and therefore a convex function, we will focus on proving that the second term, $A(\lambda)\;=\;\log\sum_{k=1}^K e^{z_k^T\lambda}$, is convex in $\lambda$. This will in turn show that each $- \log p(r_i \mid q_i; \lambda)$ in the \eqref{eq:app-convex-sum-over-dataset} sum is convex, establishing the convexity of the sum.
The convexity of $A(\lambda)$ is actually a special case of a general result about the convexity of the log-partition function in Exponential Families \citep{brown1986fundamentals}. We adapt the simple proof given in \citep{jordan2008exponential}, based on H\"older's inequality. This standard inequality states that for $p,q \in [1,\infty)$, with $\frac{1}{p} + \frac{1}{q} = 1$, and for $x_k$ and $y_k$ non-negative reals, one has
\begin{align*}
    \sum_{k=1}^K x_k y_k \leq  \left(\sum_{k=1}^K x_k^p\right)^{1/p}\, \left(\sum_{k=1}^K x_k^q\right)^{1/q}.
\end{align*}

\noindent For $\alpha\in (0,1)$, $p=\frac{1}{\alpha}, q=\frac{1}{1-\alpha}$, and for $\lambda_1,\lambda_2$ two weight vectors, we then have:
\begin{align*}
    & A(\alpha \lambda_1 + (1-\alpha) \lambda_2) \\
    &\quad = \log \sum_{k=1}^K e^{z_k^T (\alpha \lambda_1 + (1-\alpha)\lambda_2)}\\
    &\quad = \log \sum_{k=1}^K e^{\alpha z_k^T \lambda_1} \, e^{(1-\alpha)z_k^T \lambda_2} \\
    &\quad \leq \log \left[ \left(\sum_{k=1}^K (e^{\alpha z_k^T \lambda_1})^p \right)^{1/p} \left(\sum_{k=1}^K (e^{(1-\alpha) z_k^T \lambda_2})^q \right)^{1/q} \right]\\
    &\quad =\log \left[ \left(\sum_{k=1}^K e^{z_k^T \lambda_1} \right)^{1/p} \left(\sum_{k=1}^K e^{z_k^T \lambda_2}\right)^{1/q} \right]\\
    &\quad = \alpha \log \sum_{k=1}^K e^{z_k^T \lambda_1} + (1-\alpha) \log \sum_{k=1}^K e^{z_k^T \lambda_2}\\
    &\quad = \alpha A(\lambda_1) + (1-\alpha) A(\lambda_2).
\end{align*}
Thus,
\begin{equation*}
    A(\alpha \lambda_1 + (1-\alpha) \lambda_2) \leq \alpha A(\lambda_1) + (1-\alpha) A(\lambda_2),
\end{equation*}
proving the convexity of $A(\lambda)$.

\section{Dataset Construction}
\label{app:dataset}

In this appendix, we provide more details on the construction of the DnD and ELIP datasets which were introduced in Section~\ref{sec:datasets}. The contexts and responses for both datasets are in English language.

\subsection{DnD}

Our first dataset, DnD~-- named after the iconic tabletop role-playing game \textit{Dungeons and Dragons}\footnote{\url{https://dnd.wizards.com/}}~-- draws inspiration from the growing interest in role-playing language agents in recent LLM research~\citep{chen_persona_2024}, as well as past efforts in a similar setting but for different tasks~\citep{martin_dungeons_2018,callison-burch_dungeons_2022}. Simulating the persona and the actions of a character from a fantasy universe offers unique opportunities in terms of personalization due to the great variety of traits such a character can express~-- with respect to their race, class, moral alignment, background and personality.

We created the DnD dataset by proceeding according to the following pipeline, using GPT-4o for all generations:
\begin{enumerate}
    \item \textbf{Situation generation.} We started by requesting from GPT-4o to generate 200 diverse situations that a cast of players could encounter during a quest, where they would have to decide on their next action. These situations were generated within the same conversation (asking for two batches of 100 situations in successive turns) in order to ensure some diversity across situations. Table~\ref{tab:dnd-situation-actions} provides an example of the generated situations.
    \item \textbf{Situation filtering.} Although the generated situations were different, we still noted important overlap between some of them. Having two very similar situations in the training set and the test set could be seen as a ``test set leakage'' issue and affect the conclusions drawn from experiment results. For that reason, we filtered the situations using S-BERT\footnote{\url{https://sbert.net/}} as the similarity measure. Specifically, we computed the pairwise similarity between situations, and for pairs of situations with a similarity higher than 0.7, we only retained one of the two situations. This process was repeated until all pairwise similarities were lower than 0.7. After this filtering, the dataset comprised 129 situations.
    \item \textbf{Alternative actions generation.} In the next step, we generated for each of the 129 retained situations a set of 3 diverse actions as options to choose from. These actions were also generated at once to avoid having three times the same (or highly similar) action generated by the model. These possible actions along with the filtered situations define the DnD questionnaire. We show the possible actions associated to an example situation in Table~\ref{tab:dnd-situation-actions}.
    \item \textbf{Character generation.} We then proceeded to define characters who would be the ``users'' targeted by the personalization task. We requested GPT-4o to describe 10 diverse characters from the universe of Dungeons and Dragons, providing the following information for each of them: name, race, class, moral alignment, background and description. We also asked the LLM to double-check that the descriptions are fully compatible with the characters' traits, and otherwise refine the generated descriptions to make them more consistent. We show in Table~\ref{tab:dnd-character} the traits and description for the character ``Grog''.
    \item \textbf{Character pick definition.} Finally, we obtained the action picked by every character for each situation by querying GPT-4o to role-play them (based on their profile) and pick their favorite action among the set of alternative actions. The picked action is labeled as \textit{chosen} and other actions are labeled as \textit{rejected}.
\end{enumerate}
Ultimately, the DnD dataset contains 129 situations with 3 possible actions each, 10 characters, and a set of 1290 preferences corresponding to (character, situation, chosen action) tuples.

\begin{table}[t]
\centering
\begin{tabular}{ll}
\toprule
\textbf{Situation:} & \\
\multicolumn{2}{l}{
\begin{minipage}[t]{0.45\textwidth}
The lush valley you find yourself in is filled with vibrant, exotic flora and fauna. However, in the center of this paradise stands an imposing stone tower, covered in creeping vines. Smoke curls from the topmost window, and a faint smell of sulfur taints the air. The locals speak of a reclusive sorcerer who dwells within, guarding ancient secrets.
\end{minipage}
} \\
\addlinespace\addlinespace
\textbf{Action 1:} & \begin{minipage}[t]{0.32\textwidth}
I cautiously approach the stone tower, looking for any hidden traps or signs of recent activity.
\end{minipage} \\
\addlinespace
\textbf{Action 2:} & \begin{minipage}[t]{0.32\textwidth}
I call out loudly to the sorcerer, announcing my presence and asking for a parley.
\end{minipage} \\
\addlinespace
\textbf{Action 3:} & \begin{minipage}[t]{0.32\textwidth}
I decide to explore the lush valley first, hoping to find clues or hidden paths that might lead to the sorcerer's tower.
\end{minipage} \\
\bottomrule
\end{tabular}
\caption{Example of situation and alternative actions from the DnD dataset.}
\label{tab:dnd-situation-actions}
\end{table}

\begin{table}[t]
\centering
\begin{tabular}{ll}
\toprule
\textbf{Name:} & Grog Strongjaw \\
\textbf{Race:} & Half-Orc \\
\textbf{Class:} & Barbarian \\
\textbf{Moral alignment:} & Chaotic Neutral \\
\textbf{Background:} & Outlander \\
\textbf{Description:} & \\
\multicolumn{2}{l}{
\begin{minipage}[t]{0.4\textwidth}
Grog is a massive, muscle-bound half-orc with a wild mane of black hair and numerous tribal tattoos. He enjoys the thrill of battle and the freedom of the wilderness. Grog's actions are driven by his impulses and desires, often ignoring rules and societal norms.
\end{minipage}
} \\
\bottomrule
\end{tabular}
\caption{Profile generated for the character ``Grog'' from the DnD dataset.}
\label{tab:dnd-character}
\end{table}

\subsection{ELIP}

Our second dataset, ELIP (``Explain Like I Prefer''), addresses personalization in the context of a conversational assistant. Its name is a reference to the ELI5 (``Explain Like I'm 5'') dataset~\citep{fan2019eli5} from which we extracted questions. The steps we followed to form this dataset are described below:
\begin{enumerate}
    \item \textbf{Question selection.} The contexts used in ELIP correspond to questions extracted from the ELI5 dataset.\footnote{ELI-5 was distributed with a BSD license: \url{https://github.com/facebookresearch/ELI5/blob/main/LICENSE}.} We chose questions from this specific dataset rather than alternatives~-- such as those from Personalized Soups~\citep{jang2023personalized}~-- due to the open-ended nature of the ELI5 questions. This characteristic is particularly important for studying personalization, as possible ``acceptable'' responses should be sufficiently diverse to cater to diverse preferences (in terms of verbosity, tone, etc.). To further ensure that the selected questions were open-ended, diverse and of high-quality, we curated a set of 100 questions from ELI5 by manually inspecting randomly drawn questions.
    \item \textbf{Alternative responses generation.} Similarly to the procedure adopted in DnD we generated a set of diverse responses for each question using GPT-4o. Specifically, we requested the LLM to come up with 4 responses which reflect some diversity with respect to their complexity, verbosity and tone.
    \item \textbf{User definition.} The user profiles used in ELIP are based on those defined in the Personalized Soups dataset~\citep{jang2023personalized}. They correspond to the possible combinations of 3 dimensions (``Expertise'', ``Informativeness'' and ``Style'') with 2 alternative choices for each dimension, resulting in $2^3 = 8$ users. The dimension preferences are described in Table~\ref{tab:elip-dimensions}. Note that while the 4 responses generated in the previous step may reflect some combinations of these dimensions, they were not explicitly generated to satisfy a specific combination of preferences (i.e., a specific user profile).
    \item \textbf{User pick definition.} Finally, users' response picks are defined by presenting each question and its possible responses to GPT-4o, which was prompted to simulate the choice of each user based on their preferences. This resulted for each user in a \textit{chosen} response and 3 \textit{rejected} responses per question.
\end{enumerate}
The final ELIP dataset then comprises 100 situations each associated with 4 alternative responses, 8 users, and 800 preference tuples in the format (user, question, chosen response).

\begin{table*}[t]
\centering
\begin{tabular}{ll}
\toprule
\textbf{Question:} & \\
\multicolumn{2}{l}{
\begin{minipage}[t]{0.8\textwidth}
What is civil engineering?
\end{minipage}
} \\
\addlinespace\addlinespace
\textbf{Response 1:} & \begin{minipage}[t]{0.8\textwidth}
Civil engineering is the field of engineering that focuses on designing, constructing, and maintaining infrastructure like roads, bridges, buildings, and water systems to support modern society.
\end{minipage} \\
\addlinespace
\textbf{Response 2:} & \begin{minipage}[t]{0.8\textwidth}
Civil engineering is all about creating the physical structures and systems that make cities and communities function. From highways and skyscrapers to dams and sewage systems, civil engineers turn ideas into reality.
\end{minipage} \\
\addlinespace
\textbf{Response 3:} & \begin{minipage}[t]{0.8\textwidth}
Civil engineering is a professional discipline that applies principles of physics, mathematics, and materials science to the planning, analysis, and construction of infrastructure projects, ensuring safety, efficiency, and sustainability.
\end{minipage} \\
\addlinespace
\textbf{Response 4:} & \begin{minipage}[t]{0.8\textwidth}
Imagine someone who figures out how to build bridges that don’t fall down, roads that people can drive on, and tunnels that go under rivers. That’s what civil engineers do—they're like the superheroes of construction!
\end{minipage} \\
\bottomrule
\end{tabular}
\caption{Example of question and alternative responses from the ELIP dataset.}
\label{tab:elip-question-responses}
\end{table*}

\begin{table}[t]
\centering
\scalebox{0.87}{
\begin{tabular}{ll}
\toprule
\textbf{Dimension} & \textbf{Options} \\
\midrule
\textbf{Expertise} & 
\begin{minipage}[t]{0.32\textwidth}
\textbf{A:} Prefers a response that can be easily understood by a child \\
\textbf{B:} Prefers a response that is tailored for an expert audience
\end{minipage} \\
\addlinespace\addlinespace
\textbf{Informativeness} & 
\begin{minipage}[t]{0.32\textwidth}
\textbf{A:} Prefers a response that is concise and to the point, without being verbose \\
\textbf{B:} Prefers a response that is highly informative and detailed
\end{minipage} \\
\addlinespace\addlinespace
\textbf{Style} & 
\begin{minipage}[t]{0.32\textwidth}
\textbf{A:} Prefers a response that is friendly, witty, funny, and humorous \\
\textbf{B:} Prefers a response that answers in a cold and impersonal tone
\end{minipage} \\
\bottomrule
\end{tabular}
}
\caption{User preference dimensions in ELIP and their possible values. All possible combinations yield 8 user profiles: AAA, AAB, ABA, ABB, BAA, BAB, BBB.}
\label{tab:elip-dimensions}
\end{table}

\section{Experimental Setup Complements}
\label{app:exp-setup}

\subsection{Discovered Features}
\label{app:features}

Tables~\ref{tab:dnd-features} and~\ref{tab:elip-features} show the features discovered and used in FaST for DnD and ELIP, respectively. These features were obtained by generating a list of 20 features from GPT-4o\footnote{We used the gpt-4o-2024-11-20 checkpoint.} using the same domain-agnostic prompt (described in Table~\ref{tab:prompt-discovery}) based on the contexts and responses from the training set. Although our experiments are done from several train/validation/test splits, we always use the same feature set discovered from the train set of a single one of these splits. We did so as we observed only few differences in the generated features when considering different training sets. This observation was also confirmed by the preferred response prediction results on FaRM variants shown in Table~\ref{tab:rp-ablation}, which provides in (iii) the results for a feature set obtained on the training set of a different split.

\begin{table*}[t]
\centering
\scalebox{0.92}{
\begin{tabular}{lp{0.65\textwidth}}
\toprule
\textbf{Feature name} & \textbf{Description} \\
\midrule
risk\_assessment & How much does the choice involve evaluating or mitigating potential risks or dangers? \\
information\_gathering & To what extent does the choice focus on collecting information or clues? \\
direct\_action & How much does the choice involve taking immediate, decisive action? \\
caution & How cautious or careful is the approach taken in the choice? \\
problem\_solving & How much does the choice involve solving a problem or overcoming a challenge? \\
social\_interaction & To what extent does the choice involve interacting with other characters or entities? \\
magical\_investigation & How much does the choice involve the use of magic to investigate or understand the situation? \\
physical\_exertion & How much physical effort or strength is required in the choice? \\
stealth & How much does the choice rely on stealth or remaining unnoticed? \\
exploration & To what extent does the choice involve exploring or investigating the environment? \\
combat\_preparedness & How much does the choice involve preparing for or anticipating combat? \\
resourcefulness & How much does the choice involve using creativity or available resources to address the situation? \\
leadership & To what extent does the choice involve taking charge or guiding others? \\
investigative\_thoroughness & How thorough is the investigation or examination in the choice? \\
empathy & How much does the choice involve understanding or addressing the emotions or needs of others? \\
strategic\_planning & How much does the choice involve planning or strategizing for future outcomes? \\
curiosity & How much does the choice reflect a desire to learn or uncover new knowledge? \\
self\_preservation & How much does the choice prioritize personal safety or survival? \\
teamwork & To what extent does the choice involve collaboration or coordination with others? \\
mystical\_awareness & How much does the choice involve awareness or interaction with mystical or supernatural elements? \\
\bottomrule
\end{tabular}
}
\caption{Features discovered by FaST for the DnD dataset.}
\label{tab:dnd-features}
\end{table*}

\begin{table*}[t]
\centering
\scalebox{0.92}{
\begin{tabular}{lp{0.65\textwidth}}
\toprule
\textbf{Feature name} & \textbf{Description} \\
\midrule
scientific\_explanation & How much does the choice rely on scientific principles or technical details to explain the concept? \\
metaphorical\_analogy & How much does the choice use metaphors or analogies to simplify or illustrate the concept? \\
humorous\_tone & How much humor or playfulness is present in the explanation? \\
technical\_jargon & How much does the choice use specialized or technical terminology? \\
historical\_context & How much does the choice incorporate historical or cultural context into the explanation? \\
relatability & How much does the choice attempt to make the explanation relatable to everyday experiences? \\
visual\_imagery & How much does the choice evoke visual imagery to help explain the concept? \\
emotional\_engagement & How much does the choice attempt to engage the reader emotionally? \\
precision & How precise and detailed is the explanation provided in the choice? \\
practicality & How much does the choice focus on practical or real-world applications of the concept? \\
abstractness & How abstract or theoretical is the explanation in the choice? \\
biological\_focus & How much does the choice focus on biological or physiological aspects of the concept? \\
mechanical\_focus & How much does the choice focus on mechanical or physical systems in its explanation? \\
evolutionary\_perspective & How much does the choice incorporate an evolutionary perspective into the explanation? \\
philosophical\_depth & How much does the choice delve into philosophical or existential implications of the concept? \\
humancentric\_focus & How much does the choice focus on human experiences or perspectives? \\
comparative\_analysis & How much does the choice compare the concept to other related phenomena or systems? \\
educational\_tone & How much does the choice adopt an educational or instructive tone? \\
novelty & How much does the choice present the concept in a novel or unexpected way? \\
interdisciplinary\_approach & How much does the choice draw from multiple disciplines to explain the concept? \\
\bottomrule
\end{tabular}
}
\caption{Features discovered by FaST for the ELIP dataset.}
\label{tab:elip-features}
\end{table*}

\subsection{Compared Generation Approaches}

The approaches we compared in our personalized generation experiments are detailed below:
\begin{itemize}
    \item \textbf{Zeroshot} corresponds to prompting the base LLM without any fine-tuning. The prompt used in Zeroshot is shown in Table~\ref{tab:prompt-gen-zeroshot}. This approach does not take into account a specific user profile and is therefore not personalized.
    \item \textbf{RAG} (Retrieval-augmented Generation) is an in-context learning approach which includes relevant contexts in its prompt to help answer the targeted context. In this paper, we retrieve from the training set the top-$N$ most similar contexts to the targeted context. The retrieved contexts included in the prompt are provided with their associated user-chosen response. The retriever model we used is a S-BERT model\footnote{\url{https://sbert.net/}}~-- specifically, the all-MiniLM-L12-v2 model checkpoint.
    \item \textbf{Manyshot}~\citep{agarwal2024manyshot} is another in-context learning method which involves prompting an LLM with the entire set of training contexts, along with their associated user-chosen response. This approach can be applied to the \ppalli{} problem due to the limited available data for each user, making the training set small enough to fit into modern LLMs' context.
    \item \textbf{Manyshot-CoT} is an extension of Manyshot which instead proceeds in two conversation steps. First, the LLM prompted with the entire training set is requested to generate a description of the user based on the preferences they expressed in the provided contexts. Second, within the same conversation, the LLM is then asked to output a response to the targeted context based on the generated user description.
    \item \textbf{SFT} corresponds to the widely used supervised fine-tuning. For this approach, we used the pairs composed of a context and its user-chosen response as the training data.
    \item \textbf{DPO}~\citep{rafailov2023dpo}, i.e., Direct Preference Optimization, directly learns a generation policy from a preference dataset without explicitly defining a reward model. We apply this approach to the user-specific preference dataset from which we extract all possible binary preference tuples of the format (context, chosen response, rejected response).
    \item \textbf{Best-of-N} is an improved version of Zeroshot in which we instead sample $N$ responses for every context and retain the response that maximizes a given reward model (either RM or FaRM in our case).
    \item \textbf{PPO}~\citep{schulman2017ppo} corresponds to the Proximal Policy Optimization approach commonly used in reinforcement learning from human feedback~\citep{Ouyang2022}. In our experiments, we combined it with RM.\footnote{We omitted using PPO with our reward model FaRM due to the subpar performance shown by PPO in comparison to RFT and Online-DPO, and the larger memory footprint of this approach which makes it less promising for our targeted use-case of training one model per user.}
    \item \textbf{Online-DPO}~\citep{guo2024online-dpo} is a sampling-and-tuning method which iterates over the following steps: (i) sample responses from the generation policy, (ii) score the responses using a reward model (RM, or FaRM within FaST), and (iii) fine-tune the policy by applying DPO to the new preference tuples obtained from the scored responses. This corresponds to the fine-tuning approach detailed in Section~\ref{sec:finetuning}, instantiated with DPO.
    \item \textbf{RFT}~\citep{dong2023raft} is similar to Online-DPO with the only difference that the fine-tuning step is done with SFT instead of DPO.
\end{itemize}
The LLM prompted or fine-tuned in all these approaches is LLaMA-3.2-3B-Instruct (except for the experiments shown in Table~\ref{tab:gen-scores-1b} which are based on the 1B-parameter model). The version of the RM (respectively, FaRM) model used in RM w/ Best-of-N, PPO, Online-DPO, and RFT (respectively, FaST w/ Best-of-N, Online-DPO, and RFT) is the one that obtained the best results on the validation set in the preferred response prediction experiments (Table~\ref{tab:rp-main}). This corresponds to RM w/ LLaMA-3.2-3B-Instruct and FaRM w/ Phi-4-Mini-Instruct for both DnD and ELIP. 

Our experiments also included two oracle approaches in order to provide a reference on the performance that can be achieved when having access to the user profile (otherwise considered unknown and latent):
\begin{itemize}
    \item \textbf{Oracle-chosen} corresponds to simply using the response chosen by the user from the limited set of available options~-- this user-chosen response being identified during the construction of the dataset (specifically, in the \textit{user pick definition} step). This can be considered an Oracle approach as determining user-chosen responses during dataset construction is done by feeding the user profile to GPT-4o and asking it to choose the most suitable response among the alternative responses for each context.
    \item \textbf{Oracle-gen} involves prompting GPT-4o with the user profile and the targeted context to directly generate the response that is the most suitable for the user. We used GPT-4o rather than an open LLM to ensure that the generated responses have very high quality. The prompt adopted for this method is described in Table~\ref{tab:prompt-gen-oraclegen}.
\end{itemize}
Intuitively, Oracle-gen yields better (i.e., more personalized) responses than Oracle-chosen because there are cases where none of the available options associated to a question matches the user's preferences. The user then needs to pick the ``least worse'' response from these. In contrast, Oracle-gen has the freedom to generate a response entirely tailored to the user with access to their preferences, and its performance can thus be seen as an upper-bound in our task.

\subsection{Hyperparameters}
\label{app:hyperparameters}

The main hyperparameters used in our experiments for preferred response prediction and personalized generation are detailed in Table~\ref{tab:hyperparams}. For Best-of-N, RFT and Online-DPO, the same hyperparameter values were used for both FaST and its RM-based counterpart to ensure a fair comparison of their underlying preference models. For the final inference~-- i.e., the generation of the responses that are ultimately evaluated~-- we used greedy decoding by setting the temperature to 0 for all generation approaches.

\begin{table*}[t]
\centering
\scalebox{0.92}{
\begin{tabular}{ll}
\toprule
\textbf{Approach} & \textbf{Hyperparameters} \\ \midrule
\multicolumn{2}{l}{\textit{Preference models}} \\
\arrayrulecolor{lightgray}\midrule
RM & {learning\_rate = 1.41e-5}, {batch\_size = 16}, {num\_train\_epochs = 2} \\
RM-LoRA & {learning\_rate = 1.0e-4}, {batch\_size = 16}, {num\_train\_epochs = 2}, {lora\_r = 32}, {lora\_alpha = 64} \\
CPM & {learning\_rate = 0.1}, {max\_iter = 500}, {tolerance = 0.1} \\
FaRM & {learning\_rate = 0.1}, {max\_iter = 500}, {tolerance = 0.1} \\ \midrule
\multicolumn{2}{l}{\textit{Generation approaches}} \\ \midrule
RAG & {top\_n = 5} \\
SFT & {learning\_rate = 1.41e-5}, {batch\_size = 16}, {num\_train\_epochs = 10} \\
DPO & {learning\_rate = 5.0e-6}, {batch\_size = 16}, {num\_train\_epochs = 10} \\
Best-of-N & {num\_samples = 10}, {train\_temperature = 1.2} \\
RFT & {num\_samples = 10}, {train\_temperature = 1.2}, {learning\_rate = 1.41e-5}, {batch\_size = 16}, \\
 & {num\_train\_iters = 5}, {num\_train\_epochs\_per\_iter = 5} \\ 
Online-DPO & {num\_samples = 2}, {train\_temperature = 1.2}, {learning\_rate = 5.0e-6}, {batch\_size = 16}, \\
 & {num\_train\_iters = 5}, {num\_train\_epochs\_per\_iter = 5}, {beta = 0.1} \\ 
PPO & num\_train\_epochs = 100, num\_ppo\_epochs = 1, warmup\_steps = 50, num\_samples = 10, \\ 
 & missing\_eos\_penality = 1,
 kl\_coef = 0.1, learning\_rate = 1.0e-6,  batch\_size = 32 \\
\arrayrulecolor{black}\bottomrule
\end{tabular}
}
\caption{Hyperparameters adopted for each preference model and generation approach.}
\label{tab:hyperparams}
\end{table*}

\subsection{Evaluation Protocol}

This section details the evaluation methodology adopted for assessing personalized generation. We used two complementary metrics, both obtained through an LLM-judge (namely, GPT-4o-mini\footnote{We used the gpt-4o-mini-2024-07-18 checkpoint.}):

\paragraph{Score-based evaluation.}

In this evaluation, we request GPT-4o-mini to output a personalized score on a scale of 0 to 5 (higher is better). The prompt that we used is shown in Table~\ref{tab:prompt-eval-score}. This methodology and the prompt were inspired by recent score rubric-based evaluation from LLM-judges~\citep{kim2023prometheus,thonet2025elitr-bench} which showed high correlation with human judgments. As a sanity check to ensure that this methodology is meaningful and that the data enables personalization (in addition to manual checks), we verified in preliminary experiments that the scores obtained from Zeroshot, Oracle-chosen, and Oracle-gen were ordered according to the following intuitive ranking: Zeroshot < Oracle-chosen < Oracle-gen~-- which is the case on both DnD and ELIP.

\paragraph{Winrate-based evaluation.}

We also adopted a pairwise evaluation to complement the pointwise score-based evaluation described previously. Indeed, the task of grading the personalization of a response can be particularly difficult in certain cases, while it may be simpler to identify the most personalized response from a pair of responses. The prompt defined for this evaluation is provided in Table~\ref{tab:prompt-eval-winrate}. It decides on a ``winner'' between two responses for a given context and for a specific user. In case none of the response ID is returned or identified, we consider it a draw. To avoid biases related to the position of the responses (i.e., whether a response is shown first or second), we randomly swap the order of appearance when comparing two approaches. Given the high cost of running multiple pairwise comparisons (due to the use of a proprietary LLM), we restrict our winrate evaluation to the comparison between every approach and either Zeroshot or Oracle-chosen. This choice stems from the fact that Zeroshot reflects the base LLM generation and is thus a weak baseline, while Oracle-chosen is a very strong competitor for non-oracle approaches as its responses were obtained by exploiting the latent user profiles.

\subsection{Computational Resources}

The preferred response prediction experiments have been conducted using a single CPU for learning the feature weights via gradient descent in FaRM, and an A100 GPU to obtain the feature-wise scores in FaRM (pre-computed once for all users) and train RM. For the personalized generation experiments, all fine-tuning and inference have been done on a single A100 GPU, except for PPO which required being trained in a multi-GPU setting with two A100s due to its larger memory footprint.

\section{Additional Results}
\label{app:results}

\subsection{Comparing FaRM Variants on Preferred Response Prediction}
\label{app:farm}

In this section, we compare our standard version of FaRM against different variants and ablations on the preferred response prediction task, using either LLaMA-3.2-3B-Instruct or Phi-4-Mini-Instruct for feature functions. The results, shown in Table~\ref{tab:rp-ablation}, include the standard FaRM (first row) and the following variants (subsequent rows):
\begin{itemize}
    \item \textbf{FaRM w/o weighted feature scoring} computes feature-wise response scores using only the most probable score token, instead of the probability-weighted average over score tokens (see Section~\ref{sec:response-scoring} for more details). The scores in this version therefore do not account for model uncertainty unlike our standard FaRM. We observe from the table that this ablation leads to a significant degradation in the accuracy, in particular with the LLaMA-3.2-3B-Instruct backbone.
    \item \textbf{FaRM w/ logistic regression} replaces the weight learning technique described in Section~\ref{sec:farm} with logistic regression adopted in the CPM approach~\citep{go2024compositional}. This latter requires changing the preference tuples (context, chosen, rejected$_1$, \ldots, rejected$_{K-1}$) into $K-1$ pairwise preference tuples of the format (context, chosen, rejected). It can be shown that our weight learning technique reduces to logistic regression when considering pairwise preferences; therefore, the primary difference between our technique and logistic regression lies in our ability to consider $K$-way preferences. The results indicate that this ablation leads to either comparable or reduced performance relative to the standard FaRM version, thereby supporting the use of our proposed weight learning technique.
    \item \textbf{FaRM w/ score-averaged $\lambda^{(u)}$} sets the user-specific feature weight vector $\lambda^{(u)}$ directly based on the per-context feature score data, rather than learning these weights. Specifically, $\lambda^{(u)}$ is defined as the average of the feature score vectors for the responses preferred by user $u$ in the training set. Formally, $\lambda^{(u)} = \frac{1}{D} \sum_{i=1}^D \phi(q_i, r_i^{(u)})$ where $r_i^{(u)}$ is the response preferred by user $u$ for training context $q_i$ and $\phi(q_i, r_i^{(u)})$ is the vector of feature-wise scores returned by the feature functions. The results show that this simple, learning-free technique is however ineffective and leads to severe performance degradation compared to our standard FaRM.
    \item \textbf{FaRM w/ alternative feature set} corresponds to using a different set of 20 features than the ones leveraged in the standard FaRM. This alternative feature set is obtained by applying the feature discovery procedure (described in Section~\ref{sec:feature-discovery}) to the training contexts and responses from a different train/validation/test split. We include this variant to ensure a relative stability in the performance of FaRM for different discovered feature sets, which is confirmed by the results in Table~\ref{tab:rp-ablation}.
    \item \textbf{FaRM w/ \#features = 40} uses a set of 40 features rather than 20. The corresponding feature set was obtained by simply requesting 40 features instead of 20 in the prompt provided to GPT-4o in the feature discovery step (see Table~\ref{tab:prompt-discovery} for the prompt). This variant was defined to assess the impact of the number of features used and verify whether 20 is a sufficient number. We observe from the results that the performance for 20 and 40 features is comparable on the validation and test sets~-- however, with slightly increased overfitting on the training set for 40 features. This confirms that using 20 features is enough on both DnD and ELIP.
\end{itemize}

\begin{table*}[t]
\centering
\scalebox{0.85}{
\begin{tabular}{@{}lcccccccccccc@{}}
\toprule
\multirow{4}{*}{\textbf{Preference model}} & \multicolumn{6}{c}{\textbf{Acc. (\%) on DnD}} & \multicolumn{6}{c}{\textbf{Acc. (\%) on ELIP}} \\ \cmidrule(l{4pt}r{4pt}){2-7} \cmidrule(l{4pt}r{4pt}){8-13}
 & \multicolumn{3}{c}{\textbf{LLaMA-3.2-3B-It}} & \multicolumn{3}{c}{\textbf{Phi-4-Mini-It}} & \multicolumn{3}{c}{\textbf{LLaMA-3.2-3B-It}} & \multicolumn{3}{c}{\textbf{Phi-4-Mini-It}} \\ \cmidrule(l{4pt}r{4pt}){2-4} \cmidrule(l{4pt}r{4pt}){5-7} \cmidrule(l{4pt}r{4pt}){8-10} \cmidrule(l{4pt}r{4pt}){11-13}
 & \textbf{train} & \textbf{val} & \textbf{test} & \textbf{train} & \textbf{val} & \textbf{test} & \textbf{train} & \textbf{val} & \textbf{test} & \textbf{train} & \textbf{val} & \textbf{test} \\ \midrule
FaRM & 71.7 & 65.5 & 63.9 & 75.3 & 66.6 & 69.4 & 75.9 & 71.1 & 71.0 & 80.6 & 76.1 & 75.3 \\
w/o weighted feature scoring (i) & 70.6 & 58.6 & 58.4 & 74.9 & 63.6 & 67.0 & 72.6 & 64.4 & 62.0 & 79.0 & 73.1 & 72.9 \\
w/ logistic regression (ii) & 69.8 & 63.6 & 63.2 & 73.3 & 66.5 & 68.1 & 73.6 & 69.7 & 69.9 & 78.8 & 76.4 & 75.2 \\
w/ score-averaged $\lambda^{(u)}$ (iii) & 43.4 & 45.6 & 41.4 & 43.5 & 42.5 & 42.8 & 28.4 & 27.1 & 25.8 & 38.3 & 36.6 & 38.9 \\
w/ alternative feature set (iv) & 72.0 & 64.4 & 62.9 & 75.8 & 66.6 & 66.7 & 76.9 & 73.5 & 70.8 & 78.9 & 75.9 & 73.3 \\ 
w/ \#features = 40 (v) & 74.2 & 65.2 & 64.5 & 78.5 & 67.8 & 69.3 & 77.7 & 71.7 & 72.4 & 80.9 & 75.0 & 75.3 \\ \bottomrule
\end{tabular}
}
\caption{Preferred response prediction results on DnD and ELIP (higher is better), for different variants of the proposed FaRM. Specifically, \textbf{(i)} we remove the weighted feature scoring and use instead argmax scoring; \textbf{(ii)} we train FaRM with logistic regression from (chosen, rejected) pairs; \textbf{(iii)} $\lambda^{(u)}$ is set in a learning-free manner by averaging chosen responses' feature-wise score vectors; \textbf{(iv)} we adopt an alternative set of 20 features discovered from the train set of a different train/val/test split; \textbf{(v)} the requested number of features to discover is set to 40 instead of 20. Performance is measured in terms of accuracy (\%).}
\label{tab:rp-ablation}
\end{table*}

\subsection{Complementary Personalized Generation Results}
\label{app:gen-results}

In this subsection, we provide additional personalized generation results to complement those presented in Section~\ref{sec:generation-exp}:
\begin{itemize}
    \item \textbf{Full train/val/test results.} Tables~\ref{tab:gen-scores-all} and~\ref{tab:gen-wr-all} respectively provide a version of the personalized score and winrate results for the train, validation and test sets~-- rather than the average of the validation and test sets as reported in Table~\ref{tab:gen-main}. Additionally, the scores reported in Table~\ref{tab:gen-scores-all} correspond to the raw scores obtained from the LLM evaluator, rather than the improvement over the scores of the Zeroshot baseline shown in Table~\ref{tab:gen-main}.
    \item \textbf{Winrate of FaST vs RM-based variants.} Table~\ref{tab:gen-wr-add} gives the winrate results comparing FaST-based approaches with their RM-based counterparts in a pairwise manner. We can observe that in almost all cases the comparison favors FaST over RM, with the only exception of the RFT results on ELIP's test set. The superiority of FaST over RM is particularly notable on the DnD dataset.
    \item \textbf{Elo rankings.} Table~\ref{tab:elo-rankings} details the Elo rankings obtained by the different approaches. These results complement the winrates shown in Tables~\ref{tab:gen-wr-all} and~\ref{tab:gen-wr-add} by considering a more comprehensive coverage of all pairwise comparisons. The methodology we adopted to obtain these results is detailed in App.~\ref{app:elo-ranking}. Overall, we observe that FaST w/ RFT and FaST w/ Online-DPO are positioned favorably in the ranking for both DnD and ELIP. While DPO was ranked 2nd on ELIP, its poor performance on DnD makes it overall less favorable than FaST.
    \item \textbf{LoRA vs full-model fine-tuning.} Given the limited data setting of \ppalli{} as well as the one-model-per-user policy adopted in this work, we investigated the possibility of fine-tuning the generation model with low-rank adapters\footnote{For LoRA-based fine-tuning, we used the following hyperparameters: learning\_rate = 1.0e-4, lora\_r = 32, lora\_alpha = 64. Other hyperparameters were kept to their standard values reported in Table~\ref{tab:hyperparams}.} \citep{hu2022lora} instead of updating the entire model~-- with the potential advantages of limiting overfitting and reducing the GPU memory required for fine-tuning. The comparative results are reported in Table~\ref{tab:gen-scores-lora}. We can observe that LoRA-based fine-tuning overall preserves the original performance of FaST w/ RFT, showing a slight improvement on DnD alongside a slight degradation on ELIP. This suggests that using LoRA with FaST could be a viable strategy to enhance GPU memory efficiency and thus enable training on more modest devices (potentially, user devices). In contrast, RM w/ RFT experienced more pronounced performance degradation on ELIP with LoRA fine-tuning, indicating that this approach might be less robust to variations in the training method and targeted dataset.
    \item \textbf{Results for the 1B base model.} Table~\ref{tab:gen-scores-1b} shows the results for the Zeroshot baseline and the different FaST and RM-based generation approaches using LLaMA-3.2-1B-Instruct as the model being prompted or fine-tuned. These results were added to complement the results obtained for the 3B version of the model, used in all other generation experiments. These new results show the same trends as what was observed for LLaMA-3.2-3B-Instruct, with FaST-based approaches outperforming their RM-based counterparts.
\end{itemize}

\begin{table*}[t]
\centering
\scalebox{0.9}{
\begin{tabular}{llcccccc} \toprule \multicolumn{2}{c}{\multirow{2}{*}{\textbf{Approach}}} & \multicolumn{3}{c}{\textbf{Score on DnD}} & \multicolumn{3}{c}{\textbf{Score on ELIP}} \\ \cmidrule(l{4pt}r{4pt}){3-5} \cmidrule(l{4pt}r{4pt}){6-8} & & \textbf{train} & \textbf{val} & \textbf{test} & \textbf{train} & \textbf{val} & \textbf{test} \\ \midrule \multirow{2}{*}{\textit{Oracle}} & Oracle-chosen & 3.436 & 3.491 & 3.440 & 3.305 & 3.312 & 3.258 \\ & Oracle-gen & 3.916 & 3.936 & 3.924 & 3.505 & 3.483 & 3.495 \\ \arrayrulecolor{lightgray}\midrule\midrule \textit{Base} & Zeroshot & 3.053 & 3.122 & 3.065 & 2.734 & 2.673 & 2.712 \\ \midrule \multirow{3}{*}{\textit{ICL}} & RAG & 3.402 & 3.242 & 3.287 & {\ul 3.262} & 2.947 & 3.040 \\ & Manyshot & 3.408 & 3.335 & 3.324 & 3.242 & 2.943 & 3.040 \\ & Manyshot-CoT & 3.455 & 3.406 & 3.423 & 3.218 & 3.000 & 3.048 \\ \midrule \multirow{2}{*}{\textit{RM-free}} & SFT & 3.442 & 3.359 & 3.360 & 3.221 & 3.075 & 3.125 \\ & DPO & 3.426 & 3.410 & 3.421 & 3.232 & {\ul 3.210} & {\ul 3.247} \\ \midrule \multirow{4}{*}{\textit{RM}} & w/ Best-of-N & 3.273 & 3.262 & 3.239 & 2.768 & 2.798 & 2.830 \\ & w/ PPO & 3.399 & 3.392 & 3.407 & 2.784 & 2.777 & 2.790 \\ & w/ Online-DPO & 3.455 & 3.430 & 3.452 & 3.005 & 3.015 & 3.020 \\ & w/ RFT & 3.518 & 3.464 & 3.464 & 3.234 & 3.163 & 3.205 \\ \midrule \multirow{3}{*}{\textit{FaST (ours)}} & w/ Best-of-N & 3.346 & 3.351 & 3.335 & 2.874 & 2.863 & 2.900 \\ & w/ Online-DPO & {\ul 3.605} & \textbf{3.570} & \textbf{3.577} & 3.164 & 3.118 & 3.112 \\ & w/ RFT & \textbf{3.634} & {\ul 3.539} & {\ul 3.566} & \textbf{3.334} & \textbf{3.253} & \textbf{3.267} \\ \arrayrulecolor{black}\bottomrule \end{tabular}
}
\caption{Personalized generation results on DnD and ELIP, measured in terms of personalization scores (higher is better). The reported scores are the raw scores returned by the LLM-judge (unlike Table~\ref{tab:gen-main} which displayed the score improvement over the Zeroshot approach). Scores returned by the LLM-judge range from 0 to 5 (see Table~\ref{tab:prompt-eval-score} for the prompt used in the evaluation). The reported numbers result from first averaging over users (10 for DnD and 8 for ELIP), then averaging over 3 train/val/test splits. The best results (excluding oracle approaches) are shown in bold, and the second-best ones are underlined.}
\label{tab:gen-scores-all}
\end{table*}

\begin{table*}[t]
\centering
\scalebox{0.85}{
\begin{tabular}{llcccccccccccc} \toprule \multicolumn{2}{c}{\multirow{4}{*}{\textbf{Approach}}} & \multicolumn{6}{c}{\textbf{Winrate (\%) vs Zeroshot}} & \multicolumn{6}{c}{\textbf{Winrate (\%) vs Oracle-chosen}} \\ \cmidrule(l{4pt}r{4pt}){3-8} \cmidrule(l{4pt}r{4pt}){9-14} & & \multicolumn{3}{c}{\textbf{DnD}} & \multicolumn{3}{c}{\textbf{ELIP}} & \multicolumn{3}{c}{\textbf{DnD}} & \multicolumn{3}{c}{\textbf{ELIP}} \\ \cmidrule(l{4pt}r{4pt}){3-5} \cmidrule(l{4pt}r{4pt}){6-8} \cmidrule(l{4pt}r{4pt}){9-11} \cmidrule(l{4pt}r{4pt}){12-14} & & \textbf{train} & \textbf{val} & \textbf{test} & \textbf{train} & \textbf{val} & \textbf{test} & \textbf{train} & \textbf{val} & \textbf{test} & \textbf{train} & \textbf{val} & \textbf{test} \\ \midrule \multirow{2}{*}{\textit{Oracle}} & Oracle-chosen & 74.5 & 73.0 & 76.6 & 74.2 & 76.5 & 74.3 & 50.0 & 50.0 & 50.0 & 50.0 & 50.0 & 50.0 \\ & Oracle-gen & 93.5 & 94.4 & 93.6 & 96.3 & 96.2 & 95.3 & 81.0 & 81.5 & 82.0 & 88.0 & 84.3 & 86.0 \\ \arrayrulecolor{lightgray}\midrule\midrule \textit{Base} & Zeroshot & 50.0 & 50.0 & 50.0 & 50.0 & 50.0 & 50.0 & 25.5 & 27.0 & 23.4 & 25.8 & 23.5 & 25.7 \\ \midrule \multirow{3}{*}{\textit{ICL}} & RAG & 72.9 & 64.7 & 67.2 & 73.3 & 67.7 & 65.8 & 48.0 & 35.3 & 39.7 & 49.7 & 32.8 & 33.7 \\ & Manyshot & 75.8 & 72.6 & 74.6 & 72.7 & 66.2 & 66.5 & 53.4 & 46.8 & 47.2 & 47.7 & 31.8 & 33.2 \\ & Manyshot-CoT & 77.5 & 76.3 & {\ul 82.2} & 72.8 & 70.8 & 69.8 & 58.6 & 52.1 & 56.8 & 49.7 & 38.5 & 41.4 \\ \midrule \multirow{2}{*}{\textit{RM-free}} & SFT & 75.8 & 68.5 & 70.9 & 72.2 & 69.2 & 67.5 & 48.1 & 43.1 & 42.3 & 43.5 & 38.5 & 40.3 \\ & DPO & 76.5 & 73.2 & 76.5 & \textbf{76.5} & {\ul 75.3} & \textbf{76.2} & 49.6 & 48.4 & 45.9 & 52.8 & \textbf{49.7} & \textbf{55.8} \\ \midrule \multirow{4}{*}{\textit{RM}} & w/ Best-of-N & 64.0 & 62.4 & 62.5 & 52.7 & 53.2 & 53.2 & 35.7 & 32.6 & 33.7 & 29.2 & 29.0 & 31.0 \\ & w/ PPO & 71.1 & 65.8 & 71.3 & 46.9 & 49.7 & 46.7 & 44.3 & 40.3 & 40.4 & 30.1 & 31.7 & 30.2 \\ & w/ Online-DPO & 71.8 & 67.9 & 70.7 & 66.8 & 68.5 & 66.7 & 48.5 & 45.6 & 46.0 & 44.7 & 41.8 & 46.0 \\ & w/ RFT & 76.6 & 72.0 & 76.1 & 76.0 & \textbf{75.7} & {\ul 73.2} & 54.0 & 48.9 & 49.6 & {\ul 53.9} & 46.2 & 48.8 \\ \midrule \multirow{3}{*}{\textit{FaST (ours)}} & w/ Best-of-N & 73.3 & 72.2 & 72.1 & 57.3 & 56.8 & 56.3 & 41.4 & 41.2 & 38.9 & 31.0 & 28.3 & 30.0 \\ & w/ Online-DPO & {\ul 84.6} & \textbf{80.3} & \textbf{83.9} & 72.9 & 74.5 & 72.3 & {\ul 66.8} & \textbf{62.5} & \textbf{62.0} & 49.6 & {\ul 48.7} & {\ul 51.7} \\ & w/ RFT & \textbf{86.9} & {\ul 78.8} & {\ul 82.2} & {\ul 76.3} & 74.7 & 72.9 & \textbf{66.9} & {\ul 57.8} & {\ul 59.9} & \textbf{54.4} & 44.5 & 50.0 \\ \arrayrulecolor{black}\bottomrule \end{tabular}
}
\caption{Personalized generation results on DnD and ELIP, measured in terms of winrates over Zeroshot and Oracle-chosen (higher is better). The winrates have been determined by an LLM-judge (see Table~\ref{tab:prompt-eval-winrate} for the prompt used in the evaluation). The reported numbers result from first averaging over users (10 for DnD and 8 for ELIP), then averaging over 3 train/val/test splits. The best results (excluding oracle approaches) are shown in bold, and the second-best ones are underlined.}
\label{tab:gen-wr-all}
\end{table*}

\begin{table*}[t]
\centering
\scalebox{0.9}{
\begin{tabular}{lcccccc}
\toprule
\multirow{2}{*}{\textbf{Approach}} &
\multicolumn{3}{c}{\begin{tabular}{@{}c@{}}\textbf{Winrate (\%)}\\\textbf{FaST vs RM}\\\textbf{on DnD}\end{tabular}} &
\multicolumn{3}{c}{\begin{tabular}{@{}c@{}}\textbf{Winrate (\%)}\\\textbf{FaST vs RM}\\\textbf{on ELIP}\end{tabular}} \\
\cmidrule(l{4pt}r{4pt}){2-4} \cmidrule(l{4pt}r{4pt}){5-7}
& \textbf{train} & \textbf{val} & \textbf{test} & \textbf{train} & \textbf{val} & \textbf{test} \\ \midrule
Best-of-N & 60.0 & 58.8 & 58.5 & 54.2 & 53.0 & 54.5 \\
Online-DPO & 69.4 & 65.4 & 67.6 & 56.2 & 53.7 & 56.3 \\ 
RFT & 63.6 & 59.8 & 60.0 & 50.6 & 51.2 & 48.3 \\ \bottomrule
\end{tabular}
}
\caption{Winrates of FaST over RM-based generation across different generation approaches, on DnD and ELIP (higher is better). A winrate above 50\% indicates that generations from FaST are overall preferred. The winrates have been determined by an LLM-judge. The reported numbers result from first averaging over users (10 for DnD and 8 for ELIP), then averaging over 3 train/val/test splits.}
\label{tab:gen-wr-add}
\end{table*}

\begin{table*}[t]
\centering
\scalebox{0.9}{
\begin{tabular}{llr@{\hskip 2em}llr}
\toprule
\multicolumn{3}{c}{\textbf{DnD}} & \multicolumn{3}{c}{\textbf{ELIP}} \\
\cmidrule(l{2pt}r{18pt}){1-3} \cmidrule(l{-1pt}r{3pt}){4-6}
\textbf{Rank} & \textbf{Approach} & \textbf{Elo} & \textbf{Rank} & \textbf{Approach} & \textbf{Elo} \\
\midrule
1  & Oracle-gen           & 1746.4 & 1  & Oracle-gen           & 1805.1 \\
2  & FaST w/ Online-DPO   & 1682.8 & 2  & DPO                  & 1631.5 \\
3  & FaST w/ RFT          & 1627.8 & 3  & FaST w/ Online-DPO   & 1554.1 \\
4  & RM w/ RFT            & 1592.0 & 4  & RM w/ Online-DPO     & 1551.9 \\
5  & Oracle-chosen        & 1561.5 & 5  & FaST w/ RFT          & 1549.8 \\
6  & Manyshot-CoT         & 1545.2 & 6  & Oracle-chosen        & 1509.2 \\
7  & RM w/ Online-DPO     & 1539.3 & 7  & RM w/ RFT            & 1497.8 \\
8  & Manyshot             & 1514.6 & 8  & SFT                  & 1472.8 \\
9  & RM w/ PPO            & 1444.3 & 9  & Manyshot-CoT         & 1471.7 \\
10 & SFT                  & 1435.7 & 10 & RAG                  & 1444.0 \\
11 & FaST w/ Best-of-N    & 1402.9 & 11 & Manyshot             & 1432.7 \\
12 & RAG                  & 1400.0 & 12 & RM w/ PPO            & 1420.3 \\
13 & RM w/ Best-of-N      & 1383.4 & 13 & RM w/ Best-of-N      & 1420.1 \\
14 & DPO                  & 1351.6 & 14 & FaST w/ Best-of-N    & 1404.7 \\
15 & Zeroshot             & 1272.4 & 15 & Zeroshot             & 1334.3 \\
\bottomrule
\end{tabular}
}
\caption{Elo rankings obtained on the DnD and ELIP datasets (higher is better).}
\label{tab:elo-rankings}
\end{table*}

\begin{table*}[t]
\centering
\scalebox{0.9}{
\begin{tabular}{llcccccc} \toprule
\multicolumn{2}{c}{\multirow{2}{*}{\textbf{Approach}}} & \multicolumn{3}{c}{\textbf{Score on DnD}} & \multicolumn{3}{c}{\textbf{Score on ELIP}} \\
\cmidrule(l{4pt}r{4pt}){3-5} \cmidrule(l{4pt}r{4pt}){6-8}
& & \textbf{train} & \textbf{val} & \textbf{test} & \textbf{train} & \textbf{val} & \textbf{test} \\
\midrule
\textit{Base} & Zeroshot & 3.053 & 3.122 & 3.065 & 2.734 & 2.673 & 2.712 \\
\arrayrulecolor{lightgray}\midrule
\multirow{2}{*}{\textit{RM}} 
& w/ RFT (full-model fine-tuning) & 3.518 & 3.464 & 3.464 & 3.234 & 3.163 & 3.205 \\
& w/ RFT (LoRA fine-tuning) & 3.540 & 3.509 & 3.507 & 2.959 & 2.955 & 2.947 \\
\midrule
\multirow{2}{*}{\textit{FaST (ours)}} 
& w/ RFT (full-model fine-tuning) & 3.634 & 3.539 & 3.566 & 3.334 & 3.253 & 3.267 \\
& w/ RFT (LoRA fine-tuning) & 3.611 & 3.612 & 3.624 & 3.198 & 3.175 & 3.170 \\
\arrayrulecolor{black}\bottomrule
\end{tabular}
}
\caption{Comparison of the personalized generation results for full-model fine-tuning and LoRA-based fine-tuning on DnD and ELIP. Performance is measured in terms of personalization scores ranging from 0 to 5 (higher is better). The reported numbers result from first averaging over users (10 for DnD and 8 for ELIP), then averaging over 3 train/val/test splits.}
\label{tab:gen-scores-lora}
\end{table*}

\begin{table*}[t]
\centering
\scalebox{0.9}{
\begin{tabular}{llcccccc} \toprule
\multicolumn{2}{c}{\multirow{2}{*}{\textbf{Approach}}} & \multicolumn{3}{c}{\textbf{Score on DnD}} & \multicolumn{3}{c}{\textbf{Score on ELIP}} \\
\cmidrule(l{4pt}r{4pt}){3-5} \cmidrule(l{4pt}r{4pt}){6-8}
& & \textbf{train} & \textbf{val} & \textbf{test} & \textbf{train} & \textbf{val} & \textbf{test} \\
\midrule
\multirow{2}{*}{\textit{Oracle}} & Oracle-chosen & 3.436 & 3.491 & 3.440 & 3.305 & 3.312 & 3.258 \\
& Oracle-gen & 3.916 & 3.936 & 3.924 & 3.505 & 3.483 & 3.495 \\
\arrayrulecolor{lightgray}\midrule\midrule
\textit{Base} & Zeroshot & 3.063 & 3.070 & 3.079 & 2.400 & 2.333 & 2.332 \\
\midrule
\multirow{3}{*}{\textit{RM}} & w/ Best-of-N & 3.244 & 3.203 & 3.222 & 2.476 & 2.460 & 2.543 \\
& w/ Online-DPO & 3.395 & 3.394 & 3.401 & 2.835 & 2.765 & 2.897 \\
& w/ RFT & 3.440 & 3.387 & 3.406 & 2.885 & 2.758 & 2.837 \\
\midrule
\multirow{3}{*}{\textit{FaST (ours)}} & w/ Best-of-N & 3.381 & 3.359 & 3.379 & 2.603 & 2.557 & 2.650 \\
& w/ Online-DPO & \underline{3.589} & \textbf{3.575} & \textbf{3.562} & \underline{2.954} & \underline{2.852} & \underline{3.007} \\
& w/ RFT & \textbf{3.625} & \underline{3.542} & \underline{3.526} & \textbf{2.988} & \textbf{2.932} & \textbf{3.032} \\
\arrayrulecolor{black}\bottomrule
\end{tabular}
}
\caption{Personalized generation results on DnD and ELIP using LLaMA-3.2-1B-Instruct as the generation model (the oracle approaches are not affected and still rely on GPT-4o). Performance is measured in terms of personalization scores ranging from 0 to 5 (higher is better). The reported numbers result from first averaging over users (10 for DnD and 8 for ELIP), then averaging over 3 train/val/test splits. The best results (excluding oracle approaches) are shown in bold, and the second-best ones are underlined.}
\label{tab:gen-scores-1b}
\end{table*}

\subsection{Elo Ranking Methodology}
\label{app:elo-ranking}

Pairwise ranking provides a relative ordering of methods, revealing which model tends to win head-to-head comparisons. We relied on Elo, the rating system originally developed for chess \citep{elo1978rating}, as part of an evaluation pipeline in three stages:
\begin{enumerate}
    \item \textbf{Pair}: For each user/context, we generate all head-to-head matchups of two different approaches.
    \item \textbf{Play}: For each matchup, we present the two answers to an LLM-judge~-- we used GPT-4o-mini via LangChain with the same prompt that was adopted for the winrate results~-- and request a winner. We picked 5 contexts at random for each user, translating to 5 $\times$ 105 matchups (where 105 is the number of combinations of 15 approaches). We then computed the matrices of results for multiple judgment runs, %
    aka tournament, and checked that they had a very high Pearson correlation %
    despite the contexts being different~-- in practice, we observed a correlation above 0.9 on 3 runs.
    \item \textbf{Rank}: We compute the Elo scores for all the approaches from the results. We used an initial Elo of 1500 for all approaches and a $k$-factor of 16. We averaged Elo computations over 25 random shuffles of match results.
\end{enumerate}

\section{Additional Generated Samples}
\label{app:samples}

Table~\ref{tab:samples-extended} shows the responses generated by multiple approaches for a DnD and an ELIP context, as a complement to Table~\ref{tab:samples}. Personalization is done for ``Grog'' (whose profile is described in Table~\ref{tab:dnd-character}) on DnD and for ``AAA'' (whose preferences are detailed in Table~\ref{tab:elip-dimensions}) on ELIP. Additionally, to better understand the users' specific preferences and help interpret whether these are accounted for in the generated responses, Table~\ref{tab:feature-weights} reports the feature weights $\lambda^{(u)}$ learned by FaRM for these two users.

\begin{table*}[t]
\centering
\begin{minipage}{0.48\textwidth}
\centering
\begin{tabular}{lr}
\toprule
\textbf{Feature} & \textbf{Weight} \\
\midrule
risk\_assessment & \colorvaluegradient{0.3494} \\
information\_gathering & \colorvaluegradient{-0.0975} \\
direct\_action & \colorvaluegradient{0.2608} \\
caution & \colorvaluegradient{-0.1817} \\
problem\_solving & \colorvaluegradient{-0.0301} \\
social\_interaction & \colorvaluegradient{-0.0399} \\
magical\_investigation & \colorvaluegradient{-0.2242} \\
physical\_exertion & \colorvaluegradient{0.2744} \\
stealth & \colorvaluegradient{0.0470} \\
exploration & \colorvaluegradient{0.1011} \\
combat\_preparedness & \colorvaluegradient{0.2575} \\
resourcefulness & \colorvaluegradient{-0.1112} \\
leadership & \colorvaluegradient{0.0026} \\
investigative\_thoroughness & \colorvaluegradient{-0.0702} \\
empathy & \colorvaluegradient{-0.0981} \\
strategic\_planning & \colorvaluegradient{-0.0709} \\
curiosity & \colorvaluegradient{-0.0679} \\
self-preservation & \colorvaluegradient{-0.2058} \\
teamwork & \colorvaluegradient{0.0132} \\
mystical\_awareness & \colorvaluegradient{-0.0609} \\
\bottomrule
\end{tabular}
\caption*{(a) Feature weights for $u=$ ``Grog'' on DnD}
\end{minipage}
\hfill
\begin{minipage}{0.48\textwidth}
\centering
\begin{tabular}{lr}
\toprule
\textbf{Feature} & \textbf{Weight} \\
\midrule
scientific\_explanation & \colorvaluegradient{-0.1878} \\
metaphorical\_analogy & \colorvaluegradient{0.3079} \\
humorous\_tone & \colorvaluegradient{0.4055} \\
technical\_jargon & \colorvaluegradient{-0.1553} \\
historical\_context & \colorvaluegradient{-0.0134} \\
relatability & \colorvaluegradient{0.1984} \\
visual\_imagery & \colorvaluegradient{0.4033} \\
emotional\_engagement & \colorvaluegradient{0.3196} \\
precision & \colorvaluegradient{-0.2692} \\
practicality & \colorvaluegradient{-0.0998} \\
abstractness & \colorvaluegradient{-0.0020} \\
biological\_focus & \colorvaluegradient{-0.0750} \\
mechanical\_focus & \colorvaluegradient{-0.0206} \\
evolutionary\_perspective & \colorvaluegradient{0.0131} \\
philosophical\_depth & \colorvaluegradient{0.0751} \\
humancentric\_focus & \colorvaluegradient{0.1343} \\
comparative\_analysis & \colorvaluegradient{0.1283} \\
educational\_tone & \colorvaluegradient{-0.2357} \\
novelty & \colorvaluegradient{0.1881} \\
interdisciplinary\_approach & \colorvaluegradient{-0.1510} \\
\bottomrule
\end{tabular}
\caption*{(b) Feature weights for $u=$ ``AAA'' on ELIP}
\end{minipage}
\caption{Feature weights $\lambda^{(u)}$ learned by FaRM for two illustrative users: Grog on the DnD dataset \textbf{(a)} and AAA on the ELIP dataset \textbf{(b)}. The feature weights are to be interpreted in light of the users' respective profiles, detailed in Table~\ref{tab:dnd-character} for Grog and Table~\ref{tab:elip-dimensions} for AAA. Cell color intensity (red for positive weights and blue for negative weights) reflects the magnitude of the weights and thus the affinity of the user for the corresponding features. Best viewed in color.}
\label{tab:feature-weights}
\end{table*}

\begin{table*}[t]
\centering
\scalebox{0.9}{
\begin{tabular}{@{}p{4cm}p{3cm}p{3cm}@{}}
\toprule
\textbf{DnD} & & \\
\midrule
\multicolumn{3}{@{}p{13.8cm}@{}}{\textbf{Situation:} The lush valley you find yourself in is filled with vibrant, exotic flora and fauna. However, in the center of this paradise stands an imposing stone tower, covered in creeping vines. Smoke curls from the topmost window, and a faint smell of sulfur taints the air. The locals speak of a reclusive sorcerer who dwells within, guarding ancient secrets.}

\vspace{0.2cm}\\
\multicolumn{3}{@{}l@{}}{\textbf{Generated action:}} \\
\textit{Zeroshot:} & \multicolumn{2}{@{}p{9.6cm}@{}}{I draw my short sword and cautiously approach the tower's main entrance, searching for any signs of magical wards or hidden dangers.} \\
\textit{Manyshot-CoT (Grog):} & \multicolumn{2}{@{}p{9.6cm}@{}}{I carefully approach the stone tower, my hand on the hilt of my sword, and call out to the sorcerer, asking to speak with him about the ancient secrets he is said to guard.} \\
\textit{DPO (Grog):} & \multicolumn{2}{@{}p{9.6cm}@{}}{I draw my sword and begin to circle the tower, searching for a vulnerable entrance or a sign of the sorcerer's guards.} \\
\textit{RM w/ RFT (Grog):} & \multicolumn{2}{@{}p{9.6cm}@{}}{I draw my sword and climb the tower's steep stairs, ready to face the sorcerer if necessary.} \\
\textit{FaST w/ RFT (Grog):} & \multicolumn{2}{@{}p{9.6cm}@{}}{I draw my sword and charge towards the tower's entrance, ready to confront the sorcerer himself.} \\
\textit{Oracle-gen (Grog):} & \multicolumn{2}{@{}p{9.6cm}@{}}{I stride up to the tower and pound on the door, shouting for the sorcerer to come out and face me.} \\
\midrule
\midrule
\textbf{ELIP} & & \\
\midrule
\multicolumn{3}{@{}p{13.8cm}@{}}{\textbf{Question:} What is civil engineering?}

\vspace{0.2cm}\\
\multicolumn{3}{@{}l@{}}{\textbf{Generated response:}} \\
\textit{Zeroshot:} & \multicolumn{2}{@{}p{9.6cm}@{}}{Civil engineering is the application of scientific and mathematical principles to design, build, and maintain infrastructure, such as roads, bridges, buildings, and water treatment facilities, with a focus on ensuring public safety and well-being.} \\
\textit{Manyshot-CoT (AAA):} & \multicolumn{2}{@{}p{9.6cm}@{}}{Civil engineering is like being the superheroes of construction, figuring out how to build bridges that don't fall down, roads that people can drive on, and tunnels that go under rivers, all while keeping the world from crumbling around us!} \\
\textit{DPO (AAA):} & \multicolumn{2}{@{}p{9.6cm}@{}}{Imagine building the ultimate Lego castle, but instead of blocks, you're working with roads, bridges, and skyscrapers - that's basically what civil engineers do, folks!} \\
\textit{RM w/ RFT (AAA):} & \multicolumn{2}{@{}p{9.6cm}@{}}{Imagine someone who can figure out how to build a bridge, design a highway, and make sure a skyscraper doesn’t fall over? That’s a civil engineer! They’re like the superheroes of the construction world.} \\
\textit{FaST w/ RFT (AAA):} & \multicolumn{2}{@{}p{9.6cm}@{}}{Imagine planning and building the superhighways, bridges, and skyscrapers of our world—civil engineering is the team of experts who make it all happen!} \\
\textit{Oracle-gen (AAA):} & \multicolumn{2}{@{}p{9.6cm}@{}}{Civil engineering is like being a superhero for cities—designing roads, bridges, buildings, and other stuff to make sure everything works and doesn’t fall down!} \\
\bottomrule
\end{tabular}
}
\caption{Example generations from multiple approaches for $u$ = ``Grog'' (DnD) and ``AAA'' (ELIP).}
\label{tab:samples-extended}
\end{table*}

\section{Prompts}
\label{app:prompts}

We provide here the main prompts used in this paper:
\begin{itemize}
    \item \textbf{FaRM-related prompts:} Tables~\ref{tab:prompt-discovery} and~\ref{tab:prompt-scoring} describe the prompts used respectively for feature discovery (Section~\ref{sec:feature-discovery}) and response scoring with prompted LLM-based feature functions (Section~\ref{sec:response-scoring}).
    \item \textbf{Generation prompts:} Tables~\ref{tab:prompt-gen-zeroshot} and~\ref{tab:prompt-gen-oraclegen} contain the Zeroshot prompts (also used to sample candidate responses in Best-of-N, PPO, Online-DPO, and RFT) and the Oracle-gen prompts, respectively.
    \item \textbf{Evaluation prompts:} Tables~\ref{tab:prompt-eval-score} and~\ref{tab:prompt-eval-winrate} provide the prompts used to obtain a 5-point personalized score and the prompts for the winrate evaluation to compare a pair of responses, respectively.
\end{itemize}

\begin{table*}[t]
\centering
\scalebox{0.88}{
\begin{tabular}{c}
\toprule
\textit{System prompt} \\
\arrayrulecolor{lightgray}\midrule
\begin{minipage}[t]{0.95\textwidth}
\small
You are an assistant whose job is to propose a list of features that appropriately characterize the different choices proposed in contexts taken from a preference dataset.
\end{minipage}
\\
\midrule
\textit{User prompt} \\
\midrule
\begin{minipage}[t]{0.95\textwidth}
\small
You will be given a list of contexts and a list of possible choices for each of them. Your job is to suggest a set of global features that would characterize the specificities of every choice. A feature corresponds to a criterion with a score from 1 to 5 reflecting the intensity of this feature in the given choice.\\

Here is the list of contexts and their associated choices:\\

\textcolor{blue}{\{contexts\}}\\

Provide a set of \textcolor{blue}{\{num\_features\}} unique and diverse features (with no duplicate) for this list of contexts and choices, formatted in JSON following the example of the template below. Make sure that features are defined to be global and not specific to a particular subset of choices. They should also be related to the choices selected, not to a specific person selecting these choices.\\

\`{}\`{}\`{}json\\
"FEATURES": \{

\quad\quad    "\textless feature name\textgreater": \{

\quad\quad\quad\quad        "attribute\_desc": "\textless a question which describes the feature and whose answer should assess the intensity of the feature in a given choice\textgreater",

\quad\quad\quad\quad        "attr\_min": "\textless a lowercase participle phrase describing what is expected for the lowest intensity of the feature\textgreater",

\quad\quad\quad\quad        "attr\_max": "\textless a lowercase participle phrase describing what is expected for the highest intensity of the feature\textgreater"

\quad\quad    \},

\quad\quad    ...\\
\}\\
\`{}\`{}\`{}
\end{minipage}
\\
\arrayrulecolor{black}\bottomrule
\end{tabular}
}
\caption{Feature discovery prompt. This prompt is generic with respect to the nature and task of the preference dataset, and was used for generating features for both DnD and ELIP. It takes as input all the training contexts and their possible responses (\textcolor{blue}{contexts}) and the number of features to generate (\textcolor{blue}{num\_features}). The features discovered for DnD and ELIP are shown in Tables~\ref{tab:dnd-features} and~\ref{tab:elip-features}, respectively.}
\label{tab:prompt-discovery}
\end{table*}

\begin{table*}[t]
\centering
\scalebox{0.88}{
\begin{tabular}{c@{\hskip 0.04\textwidth}c}
\toprule
\textbf{DnD} & \textbf{ELIP} \\
\midrule
\multicolumn{2}{c}{\textit{System prompt}}  \\
\arrayrulecolor{lightgray}\midrule
\begin{minipage}[t]{0.48\textwidth}
\small
You are a scoring assistant that evaluates player character actions in a game of Dungeons \& Dragons.
\end{minipage}
&
\begin{minipage}[t]{0.48\textwidth}
\small
You are a scoring assistant that evaluates responses generated by an AI assistant.
\end{minipage}
\\
\midrule
\multicolumn{2}{c}{\textit{User prompt}} \\
\midrule
\begin{minipage}[t]{0.48\textwidth}
\small
You will be given a situation encountered in a game of Dungeons \& Dragons, and an action that a player character may choose in this situation. Your job is to rate the action based on the following criterion: \textcolor{blue}{\{attribute\_desc\}}. Score the action on a scale from 1 to 5 where 1 means \textcolor{blue}{\{attr\_min\}} and 5 means \textcolor{blue}{\{attr\_max\}}.

\vspace{1em}
Here are the situation and the related action:

\vspace{1em}
\# Situation: \textcolor{blue}{\{context\}} \\
\# Action: \textcolor{blue}{\{response\}}

\vspace{1em}
Answer by outputting a number from 1 to 5 (and nothing else).

\vspace{1em}
Score:
\end{minipage}
&
\begin{minipage}[t]{0.48\textwidth}
\small
You will be given a question that can be submitted to an AI assistant, and a response that attempts to answer this question. Your job is to rate the response based on the following criterion: \textcolor{blue}{\{attribute\_desc\}}. Score the response on a scale from 1 to 5 where 1 means \textcolor{blue}{\{attr\_min\}} and 5 means \textcolor{blue}{\{attr\_max\}}. Here are the question and the related response:

\vspace{1em}
\# Question: \textcolor{blue}{\{context\}} \\
\# Response: \textcolor{blue}{\{response\}}

\vspace{1em}
Answer by outputting a number from 1 to 5 (and nothing else).

\vspace{1em}
Score:
\end{minipage}
\\
\arrayrulecolor{black}\bottomrule
\end{tabular}
}
\caption{Feature function prompts for DnD (left) and ELIP (right). These prompts are used to obtain feature-wise response scores. The feature is specified by the fields \textcolor{blue}{attribute\_desc} (overall description of the feature), \textcolor{blue}{attr\_min} (description of the minimum score) and \textcolor{blue}{attr\_max} (description of the maximum score) generated in the feature discovery step.}
\label{tab:prompt-scoring}
\end{table*}

\begin{table*}[t]
\centering
\scalebox{0.9}{
\begin{tabular}{c@{\hskip 0.04\textwidth}c}
\toprule
\textbf{DnD} & \textbf{ELIP} \\
\midrule
\multicolumn{2}{c}{\textit{System prompt}} \\
\arrayrulecolor{lightgray}\midrule
\begin{minipage}[t]{0.48\textwidth}
\small
You are role-playing a character in a game of Dungeons \& Dragons.
\end{minipage}
&
\begin{minipage}[t]{0.48\textwidth}
\small
You are an AI assistant that write responses to answer user questions.
\end{minipage}
\\
\midrule
\multicolumn{2}{c}{\textit{User prompt}} \\
\midrule
\begin{minipage}[t]{0.48\textwidth}
\small
You will be given a situation encountered in a game of Dungeons \& Dragons. Your job is to write the action your character would choose for this situation. The action should be a single sentence describing your character's immediate action in the first person. Keep it concise, focusing only on your character's intent without describing the environment, emotions, or potential consequences.

\vspace{1em}
Here is the situation:

\vspace{1em}
\# Situation: \textcolor{blue}{\{context\}} \\
\# Chosen action:
\end{minipage}
&
\begin{minipage}[t]{0.48\textwidth}
\small
You will be given a question. Your job is to write a response using the tone and style of your choice. The length of your response can range from a single sentence to a short paragraph. Do not include any introduction, preamble, explanation or conclusion — only the direct response to the question.

\vspace{1em}
Here is the question:

\vspace{1em}
\# Question: \textcolor{blue}{\{context\}} \\
\# Response:
\end{minipage}
\\
\arrayrulecolor{black}\bottomrule
\end{tabular}
}
\caption{Zeroshot generation prompts for DnD (left) and ELIP (right). These prompts are used to generate responses in the Zeroshot approach and sample candidate responses for Best-of-N, PPO, Online-DPO, and RFT.}
\label{tab:prompt-gen-zeroshot}
\end{table*}

\begin{table*}[t]
\centering
\scalebox{0.9}{
\begin{tabular}{c@{\hskip 0.04\textwidth}c}
\toprule
\textbf{DnD} & \textbf{ELIP} \\
\midrule
\multicolumn{2}{c}{\textit{System prompt}}  \\
\arrayrulecolor{lightgray}\midrule
\begin{minipage}[t]{0.48\textwidth}
\small
You are role-playing a character in a game of Dungeons \& Dragons.
\end{minipage}
&
\begin{minipage}[t]{0.48\textwidth}
\small
You are an AI assistant that write responses to answer user questions.
\end{minipage}
\\
\midrule
\multicolumn{2}{c}{\textit{User prompt}} \\
\midrule
\begin{minipage}[t]{0.48\textwidth}
\small
You will be given your character's description (including their race, class, moral alignment, background and personality) and a situation encountered in a game of Dungeons \& Dragons. Your job is to write the action your character would choose for this situation. The action should be a single sentence describing your character's immediate action in the first person. Keep it concise, focusing only on your character's intent without describing the environment, emotions, or potential consequences.

\vspace{1em}
Here is your character's description:

\vspace{1em}
\textcolor{blue}{\{profile\_desc\}}

\vspace{1em}
Here is the situation:

\vspace{1em}
\# Situation: \textcolor{blue}{\{context\}} \\
\# Chosen action:
\end{minipage}
&
\begin{minipage}[t]{0.48\textwidth}
\small
You will be given a description of the preferences of a user (including their expertise, preferred informativeness level, and preferred style) and a question. Your job is to write a response to the question while aligning as closely as possible with the user's preferences. The length of your response can range from a single sentence to a short paragraph. Do not include any introduction, preamble, explanation or conclusion — only the direct response to the question.

\vspace{1em}
Here is the description of the user's preferences:

\vspace{1em}
\textcolor{blue}{\{profile\_desc\}}

\vspace{1em}
Here is the question:

\vspace{1em}
\# Question: \textcolor{blue}{\{context\}} \\
\# Response:
\end{minipage}
\\
\arrayrulecolor{black}\bottomrule
\end{tabular}
}
\caption{Oracle-gen generation prompts for DnD (left) and ELIP (right). The profile of the user is indicated to Oracle-gen via the field \textcolor{blue}{profile\_desc}.}
\label{tab:prompt-gen-oraclegen}
\end{table*}

\begin{table*}[t]
\centering
\scalebox{0.9}{
\begin{tabular}{c@{\hskip 0.04\textwidth}c}
\toprule
\textbf{DnD} & \textbf{ELIP} \\
\midrule
\multicolumn{2}{c}{\textit{System prompt}}  \\
\arrayrulecolor{lightgray}\midrule
\begin{minipage}[t]{0.48\textwidth}
\small
You are an evaluator whose task is to evaluate the appropriateness of a character's action in the context of a situation from a fantasy role-playing game like Dungeons \& Dragons.
\end{minipage}
&
\begin{minipage}[t]{0.48\textwidth}
\small
You are an evaluator whose task is to evaluate how appropriate the response of an AI assistant is for a given user in the context of a specific question submitted to the AI assistant.
\end{minipage}
\\
\midrule
\multicolumn{2}{c}{\textit{User prompt}} \\
\midrule
\begin{minipage}[t]{0.48\textwidth}
\small
\#\#\# Task description:

You are provided below with a character's description (including their race, class, moral alignment, background and personality), a situation, an action to evaluate, and a score rubric representing evaluation criteria.

Write a detailed feedback as follows: \\ 
1. Start by listing the character's traits and expected skills which are reflected in the action, and those which are not reflected in it. \\ 
2. Based on these elements, write a detailed feedback that assesses to what extent the action is tailored to the character by strictly following the given score rubric. \\ 
3. After writing the feedback, write a score that is an integer between 0 and 5. You should refer to the score rubric. \\ 
4. The output format should first include the feedback and then indicate the integer score in \textbackslash boxed\{\}. \\ 
5. Please do not write any other opening, closing, and explanations.

\vspace{1em}
\#\#\# Character's description: \\
\textcolor{blue}{\{profile\_desc\}}

\vspace{1em}
\#\#\# Situation: \\
\textcolor{blue}{\{context\}}

\vspace{1em}
\#\#\# Action to evaluate: \\
\textcolor{blue}{\{generated\_response\}}

\vspace{1em}
\#\#\# Score rubric:

[Evaluate to what extent the action reflects the described character's traits (including their race, class, moral alignment, background and personality) and implicitly expected skills.]\\
Score 0: The action is irrelevant to the given situation or inconsistent with the D\&D rules and setting.\\
Score 1: The action is unlikely or contradictory for the character, given their traits and expected skills.\\
Score 2: The action is generic and could be performed by any D\&D character regardless of their traits and expected skills.\\
Score 3: The action reflects some of the character's traits and expected skills, but important aspects of their description are overlooked.\\
Score 4: The action aligns well with most of the character's traits and expected skills, though there is room for slight improvement in tailoring.\\
Score 5: The action is precisely tailored to the character, demonstrating a comprehensive alignment with their traits and expected skills.

\vspace{1em}
\#\#\# Feedback:
\end{minipage}
&
\begin{minipage}[t]{0.48\textwidth}
\small
\#\#\# Task description:

You are provided below with a description of a user's preferences (expressed along three dimensions -- required expertise, informativeness level, and style), a question, a response to evaluate, and a score rubric representing evaluation criteria.

Write a detailed feedback as follows: \\ 
1. Start by listing the user's preferences which are satisfied in the response, the user's preferences which are ignored, and the user's preferences which are contradicted. \\
2. Based on these elements, write a detailed feedback that assesses to what extent the response is tailored to the user by strictly following the given score rubric. \\ 
3. After writing the feedback, write a score that is an integer between 0 and 5. You should refer to the score rubric. \\ 
4. The output format should first include the feedback and then indicate the integer score in \textbackslash boxed\{\}. \\ 
5. Please do not write any other opening, closing, and explanations.

\vspace{1em}
\#\#\# User's description: \\
\textcolor{blue}{\{profile\_desc\}}

\vspace{1em}
\#\#\# Question: \\
\textcolor{blue}{\{context\}}

\vspace{1em}
\#\#\# Response to evaluate: \\
\textcolor{blue}{\{generated\_response\}}

\vspace{1em}
\#\#\# Score rubric:

[Evaluate to what extent the response reflects the described user's preferences (including their expertise, informativeness requirements, and preferred style).]\\
Score 0: The response is irrelevant to the given question.\\
Score 1: The response overall contradicts the user's preferences (i.e., two or more preference dimensions are contradicted).\\
Score 2: The response ignores the user's preferences (i.e., all three preference dimensions are ignored, or one preference dimension is contradicted).\\
Score 3: The response reflects some of the user's preferences (i.e., one preference dimension is satisfied, and no preference dimension is contradicted).\\
Score 4: The response aligns well with most of the user's preferences (i.e., two preference dimensions are satisfied, and no preference dimension is contradicted).\\
Score 5: The response is precisely tailored to the user (i.e., all three preference dimensions are satisfied).

\vspace{1em}
\#\#\# Feedback:
\end{minipage}
\\
\arrayrulecolor{black}\bottomrule
\end{tabular}
}
\caption{Score-based evaluation prompts for DnD (left) and ELIP (right). These prompts follow the score rubric-based LLM-judge methodology inspired from~\citet{kim2023prometheus,thonet2025elitr-bench}. The 0-5 output scores assess the level of personalization of the response (\textcolor{blue}{generated\_response}) based on the user's preferences (\textcolor{blue}{profile\_desc}) according to the specified score rubric.}
\label{tab:prompt-eval-score}
\end{table*}

\begin{table*}[t]
\centering
\scalebox{0.9}{
\begin{tabular}{c@{\hskip 0.04\textwidth}c}
\toprule
\textbf{DnD} & \textbf{ELIP} \\
\midrule
\multicolumn{2}{c}{\textit{System prompt}} \\
\arrayrulecolor{lightgray}\midrule
\begin{minipage}[t]{0.48\textwidth}
\small
You are an evaluator whose task is to determine the most appropriate action for a character among two given choices, in the context of a situation from a fantasy role-playing game like Dungeons \& Dragons.
\end{minipage}
&
\begin{minipage}[t]{0.48\textwidth}
\small
You are an evaluator whose task is to determine the most appropriate response for a user among two given choices, in the context of a question to an AI assistant.
\end{minipage}
\\
\midrule
\multicolumn{2}{c}{\textit{User prompt}} \\
\midrule
\begin{minipage}[t]{0.48\textwidth}
\small
\#\#\# Task description:

You are provided below with a character's description (including their race, class, moral alignment, background and personality), a situation, and two actions to compare (with IDs 1 and 2).\\
Write a detailed feedback as follows: \\
1. Write a detailed feedback that assesses to what extent each of the two actions is tailored to the character's traits and expected skills based on their description.\\
2. After writing the feedback, write the ID of the action (1 or 2) which is the most suitable for the character.\\
3. The output format should first include the feedback and then indicate the ID in \textbackslash boxed\{\}.\\
4. Please do not write any other opening, closing, and explanations.

\vspace{1em}
\#\#\# Character's description:\\
\textcolor{blue}{\{profile\_desc\}}

\vspace{1em}
\#\#\# Situation:\\
\textcolor{blue}{\{context\}}

\vspace{1em}
\#\#\# Action 1:\\
\textcolor{blue}{\{generated\_response1\}}

\vspace{1em}
\#\#\# Action 2:\\
\textcolor{blue}{\{generated\_response2\}}

\vspace{1em}
\#\#\# Feedback:
\end{minipage}
&
\begin{minipage}[t]{0.48\textwidth}
\small
\#\#\# Task description:

You are provided below with a user's description (including their expertise, preferred informativeness level, and preferred style), a question, and two responses to compare (with IDs 1 and 2).\\
Write a detailed feedback as follows: \\
1. Write a detailed feedback that assesses to what extent each of the two responses is tailored to the user's preferences based on their description.\\
2. After writing the feedback, write the ID of the action (1 or 2) which is the most suitable for the user.\\
3. The output format should first include the feedback and then indicate the ID in \textbackslash boxed\{\}.\\
4. Please do not write any other opening, closing, and explanations.

\vspace{1em}
\#\#\# User's description:\\
\textcolor{blue}{\{profile\_desc\}}

\vspace{1em}
\#\#\# Question:\\
\textcolor{blue}{\{context\}}

\vspace{1em}
\#\#\# Response 1:\\
\textcolor{blue}{\{generated\_response1\}}

\vspace{1em}
\#\#\# Response 2:\\
\textcolor{blue}{\{generated\_response2\}}

\vspace{1em}
\#\#\# Feedback:
\end{minipage}
\\
\arrayrulecolor{black}\bottomrule
\end{tabular}
}
\caption{Winrate-based evaluation prompts for DnD (left) and ELIP (right). These prompts enable pairwise evaluation of two responses (\textcolor{blue}{generated\_response1} and \textcolor{blue}{generated\_response2}) generated by a pair of approaches, picking the response that best fits the user's preferences (\textcolor{blue}{profile\_desc}).}
\label{tab:prompt-eval-winrate}
\end{table*}

\end{document}